\documentclass[10pt,twocolumn,letterpaper]{article}

\usepackage{cvpr}              % To produce the CAMERA-READY version
\usepackage{graphicx}
\usepackage{amsmath}
\usepackage{amssymb}
\usepackage{booktabs}
\usepackage{makecell,rotating}
\usepackage{bbding}
\usepackage{multirow}
\usepackage{tabularx}
\usepackage{colortbl}
\usepackage{color}
\definecolor{mygray}{gray}{.9}
\definecolor{myred}{rgb}{0.698,0.133,0.133}
\usepackage{sidecap}
\usepackage{enumerate}
\usepackage{makecell}
\usepackage{hhline}
\usepackage{enumitem}
\usepackage[misc]{ifsym}
\usepackage[ruled,linesnumbered]{algorithm2e}
\newtheorem{assumption}{Assumption}[section]
\usepackage[pagebackref,breaklinks,colorlinks]{hyperref}

\usepackage[capitalize]{cleveref}
\crefname{section}{Sec.}{Secs.}
\Crefname{section}{Section}{Sections}
\Crefname{table}{Table}{Tables}
\crefname{table}{Tab.}{Tabs.}

\def\cvprPaperID{xxxx} % *** Enter the CVPR Paper ID here
\def\confName{CVPR}
\def\confYear{2022}

\begin{document}

\title{Slimmable Domain Adaptation}

\author{Rang Meng$^2$, Weijie Chen$^{1,2,}$\footnotemark[2], Shicai Yang$^2$, Jie Song$^1$, Luojun Lin$^3$, Di Xie$^2$\\
Shiliang Pu$^2$, Xinchao Wang$^4$, Mingli Song$^1$, Yueting Zhuang$^{1,}$\footnotemark[2]\\
{\normalsize $^1$Zhejiang University, $^2$Hikvision Research Institute, $^3$Fuzhou University, $^4$National University of Singapore}\\
{\tt\small \{mengrang, chenweijie5,yangshicai,xiedi,pushiliang.hri\}@hikvision.com} \\
{\tt\small \{sjie,songml,yzhuang\}@zju.edu.cn, linluojun2009@126.com, xinchao@nus.edu.sg}
}

\maketitle

\renewcommand{\thefootnote}{\fnsymbol{footnote}}
\footnotetext[2]{Corresponding author}

%%%%%%%%% ABSTRACT
\begin{abstract}
Vanilla unsupervised domain adaptation methods tend to optimize the model with fixed neural architecture, which is not very practical in real-world scenarios since the target data is usually processed by different resource-limited devices. It is therefore of great necessity to facilitate architecture adaptation across various devices. In this paper, we introduce a simple framework, Slimmable Domain Adaptation, to improve cross-domain generalization with a weight-sharing model bank, from which models of different capacities can be sampled to accommodate different accuracy-efficiency trade-offs. The main challenge in this framework lies in simultaneously boosting the adaptation performance of numerous models in the model bank. To tackle this problem, we develop a Stochastic EnsEmble Distillation method to fully exploit the complementary knowledge in the model bank for inter-model interaction. Nevertheless, considering the optimization conflict between inter-model interaction and intra-model adaptation, we augment the existing bi-classifier domain confusion architecture into an Optimization-Separated Tri-Classifier counterpart. After optimizing the model bank, architecture adaptation is leveraged via our proposed Unsupervised Performance Evaluation Metric. Under various resource constraints, our framework surpasses other competing approaches by a very large margin on multiple benchmarks.  It is also worth emphasizing that our framework can preserve the performance improvement against the source-only model even when the computing complexity is reduced to $1/64$. Code will be available at \url{https://github.com/hikvision-research/SlimDA}.
\end{abstract}

%%%%%%%%% BODY TEXT
%---------------------------------introduction---------------------------------

\section{Introduction}
Deep neural networks are usually trained on the offline-collected images (labeled source data) and then embedded in edge devices to test the images sampled from new scenarios (unlabeled target data). This paradigm, in practice, degrades the network performance due to the domain shift. Recently, more and more researchers have delved into unsupervised domain adaptation (UDA) to address this problem.

\begin{figure}[t]
\begin{center}
\includegraphics[width=0.99\linewidth]{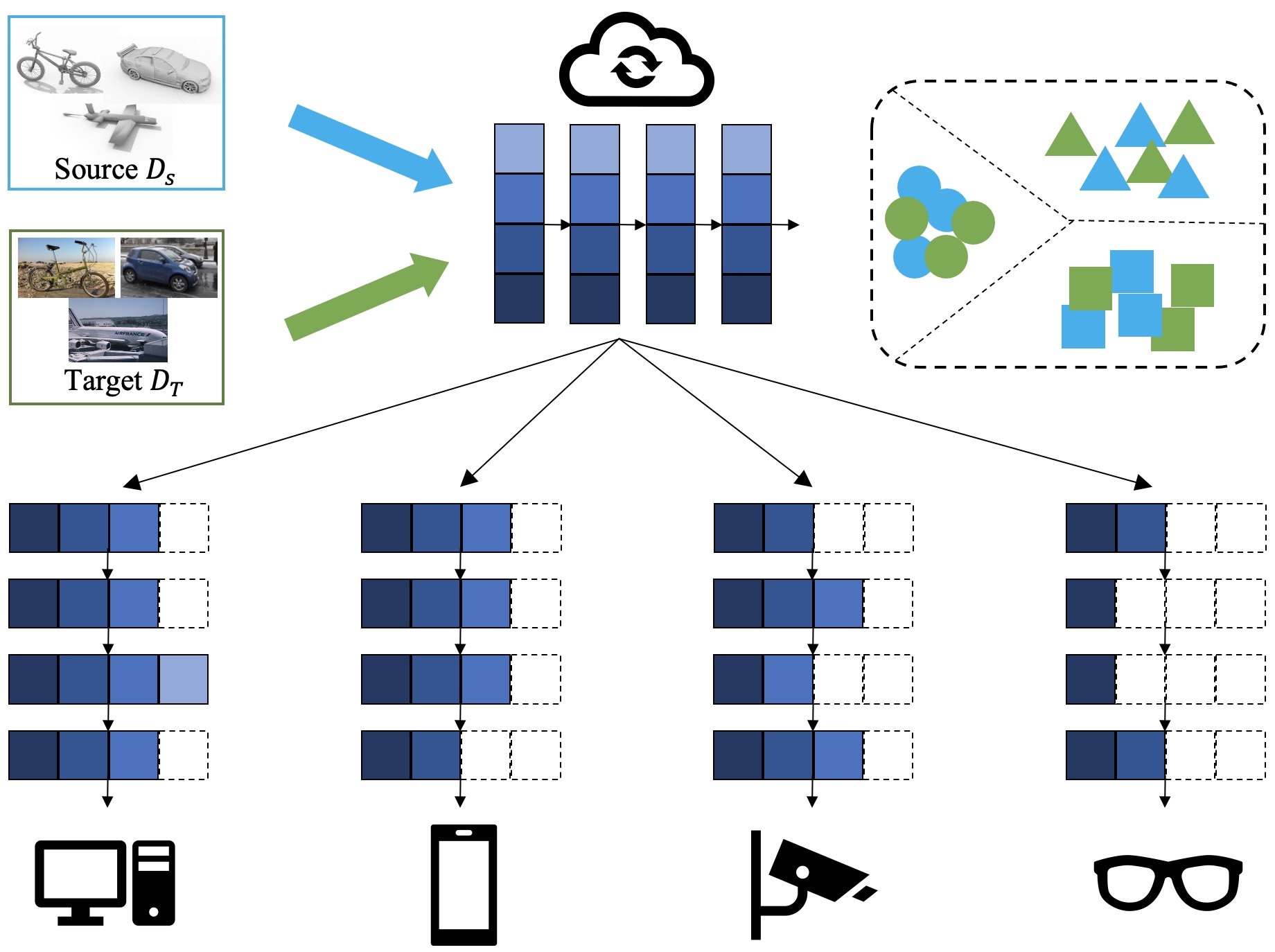}
\end{center}
   \vspace{-1.5em}
   \caption{SlimDA: We only adapt once on cloud computing center but can flexibly sample models with diverse capacities to distribute to different resource-limited edge devices.}
   \label{motivation}
   \vspace{-2em}
\end{figure}

Vanilla UDA aims to align source data and target data into a joint representation space so that the model trained on source data can be well generalized to target data \cite{MMD,GRL,ADDA,MCD,STAR,2020A,2021Self}. Unfortunately, there is still a gap between academic studies and industrial needs: most existing UDA methods only perform weight adaptation with fixed neural architecture yet cannot fit the requirements of various devices in the real-world applications efficiently. Taking the example of a widely-used application scenario as shown in Fig.\ref{motivation}, a domain adaptive model trained on a powerful cloud computing center is urged to be distributed to different resource-limited edge devices like laptops, smart mobile phones, and smartwatches, for real-time processing. 
In this scenario, vanilla UDA methods have to train a series of models with different capacities and architectures time-and-again to fit the requirements of devices with different computation budgets, which is expensive and time-consuming. 
To remedy the aforementioned issue, we propose Slimmable Domain Adaptation (SlimDA), in which we only train our model once so that the customized models with different capacities and architectures can be sampled flexibly from it to supply the demand of devices with different computation budgets. 

Although slimmable neural networks \cite{2018Slimmable,0Universally,Yu2019NetworkSB} had been studied in the supervised tasks, in which models with different layer widths (\emph{i.e.}, channel number) can be coupled into a weight-sharing model bank for optimization,
there remain two challenges when slimmable neural networks meet unsupervised domain adaptation: 1) \emph{Weight adaptation}: How to simultaneously boost the adaptation performance of all models in the model bank? 2) \emph{Architecture adaptation}: Given a specific computational budget, how to search an appropriate model on the unlabeled target data?

\renewcommand\arraystretch{1} 
\begin{table}[t]
\tiny
	\centering
	\resizebox{.48\textwidth}{!}{
	\begin{tabular}{ccccc}
	\hline
	Methods & Data Type & Teacher & Student & Model\\
	\midrule
	CKD & Labeled & Single & Single & Fixed\\
	\midrule
	SEED& Unlabeled & Multiple & Multiple & Stochastic\\
	\hline
	\end{tabular}
	}
	\vskip -0.05in
	\caption{Conventional Knowledge Distillation (CKD) vs. Stochastic EnsEmble Distillation (SEED).}
	\label{SEED}
	\vskip -0.1in
\end{table}

For the first challenge, there is a straightforward baseline in which UDA methods are directly applied to each model sampled from the model bank. However, this paradigm neglects to exploit the complementary knowledge 
among tremendous neural architectures in the model bank.
To remedy this issue, we propose Stochastic EnsEmble Distillation (SEED) to 
interact the models in the model bank so as to suppress the uncertainty of intra-model adaptation on the unlabeled target data. 
SEED is a curriculum mutual learning framework in which the expectation of the predictions from stochastically-sampled models are exploited to assist domain adaptation of the model bank. The differences between SEED and the conventional knowledge distillation are shown in Table \ref{SEED}.
As for intra-model adaptation, we borrow the solution from the state-of-the-art bi-classifier-based domain confusion method (such as SymNet~\cite{SymNet} and MCD~\cite{MCD}). Nevertheless, we analyze that there exists an optimization conflict between inter-model interaction and intra-model adaptation, which motivates us to augment an Optimization-Separated Tri-Classifier (OSTC) to modulate the optimization between them.

For the second challenge, it is intuitive to search models with optimal adaptation performance under different computational budgets after training the model bank.
However, unlike performance evaluation in the supervised tasks, none of the labeled target data is available. 
To be compatible with the unlabeled target data, we exploit the model with the largest capacity as an anchor to guide performance ranking in the model bank, since the larger models tend to be more accurate as empirically proven in \cite{pmlr-v97-ying19a}.
In this way, we propose an Unsupervised Performance Evaluation Metric which is eased into the output discrepancy between the candidate model and the anchor model.
The smaller the metric is, the better the performance is assumed to be.

Extensive ablation studies and experiments are carried out on three popular UDA benchmarks, \emph{i.e.}, ImageCLEF-DA \cite{long2017deep}, Office-31 \cite{2010Adapting}, and Office-Home \cite{Venkateswara48}, which demonstrate the effectiveness of the proposed framework. Our method can achieve state-of-the-art results compared with other competing methods. It is worth emphasizing that our method can preserve the performance improvement against the source-only model even when the computing complexity is reduced to $1/64\times$. To summarize, our main contributions are listed as follows:
\begin{itemize}[leftmargin=12pt, topsep=2pt, itemsep=0pt]
\item We propose SlimDA, a ``once-for-all'' framework to jointly accommodate the adaptation performance and the computation budgets for resource-limited devices.
\item We propose SEED to simultaneously boost the adaptation performance of all models in the model bank. In particular, we design an Optimization-Separated Tri-Classifier to modulate the optimization between intra-model adaptation and inter-model interaction.
\item We propose an Unsupervised Performance Evaluation Metric to facilitate architecture adaptation.
\item Extensive experiments verify the effectiveness of our proposed SlimDA framework, which can surpass other state-of-the-art methods by a large margin.
\end{itemize}

%-------------------------------------------------------------------------
\section{Related Work}

\subsection{Unsupervised Domain Adaptation}
Existing UDA methods aim to improve the model performance on the unlabeled target domain. In the past few years, discrepancy-based methods \cite{MMD,Deepcoral,AssocDA} and adversarial optimization methods \cite{MCD,CoGAN,PixelDA,CYCADA,GRL} are proposed to solve this problem via domain alignment. Specifically, SymNet \cite{SymNet} develops a bi-classifier architecture to facilitate category-level domain confusion. Recently, Li \emph{et.al.} \cite{Li2020NetworkAS} attempts to learn optimal architectures to further boost the performance on the target domain, which proves the significance of network architecture for UDA. These UDA methods focus on achieving a specific model with
 better performance on the target domain.

\subsection{Neural Architecture Search}
Neural Architecture Search (NAS) methods aim to search for optimal architectures automatically through reinforcement learning \cite{zoph2016neural,cai2018path,zoph2018learning,tan2019efficientnet,tan2019mnasnet}, evolution methods \cite{liu2017hierarchical,real2019regularized,perez2019mfas,elsken2018efficient}, gradient-based methods \cite{shin2018differentiable,luo2018neural,liu2019auto,wu2019fbnet,xie2018snas} and so on. Recently, one-shot methods \cite{Yu2019NetworkSB,2020Neural,Guo2020SinglePO,Cai2020OnceFA,xie2018snas,brock2017smash} are very popular since only one super-network is required to train, and numerous weight-sharing sub-networks of various architectures are optimized simultaneously. In this way, the optimal network architecture can be searched from the model bank. In this paper, we highlight that UDA is an unnoticed yet significant scenario for NAS, since they can be cooperated to optimize a scene-specific lightweight architecture in an unsupervised way.

\subsection{Cross-domain Network Compression}
Chen \emph{et.al.} \cite{ijcai2019-291} proposes a cross-domain unstructured pruning method. Yu \emph{et.al.} \cite{Yu2019AcceleratingDU} adopts MMD \cite{MMD} to minimize domain discrepancy and prunes filters in a Taylor-based strategy, and
Yang~\emph{et.al.}~\cite{Yang2020Distill,Yang2020Factor} focuses on
compressing graph neural networks.
Feng \emph{et.al.} \cite{Feng2020ADMPAA}  conducts adversarial training between the channel-pruned network and the full-size network. However, the performances of the existing methods still have a great improvement space. Moreover, their methods are not flexible enough to obtain numerous optimal models under diverse resource constraints.
%-------------------------------------------------------------------------

\section{Preliminary}

\subsection{Bi-Classifier Based Domain Confusion}
\subsubsection{Notation}
A labeled source data $\mathcal{D}_s=\{(x_i^s,y_i^s)\}_{i=1}^{n_s}$ and an unlabeled target data $\mathcal{D}_t=\{(x_i^t)\}_{i=1}^{n_t}$ are provided for training. SymNet \cite{SymNet} is composed of a feature extractor $F$ and two task classifiers $C^s$ and $C^t$. A novel design in SymNet is to construct a new classifier $C^{st}$ which shares the neurons with $C^s$ and $C^t$. $C^{st}$ is designed for domain discrimination and domain confusion without an explicit domain discriminator. The probability outputs of $C^s$, $C^t$ and $C^{st}$ are $\mathbf{g}(x
;F,C^s)\in[0,1]^K$, $\mathbf{g}(x;F,C^t)\in[0,1]^K$ and $\mathbf{g}(x;F,C^{st})\in[0,1]^{2K}$ respectively, where $K$ is the class number of the task. The $k^{th}$ element of the probability output can be written as $\mathbf{g}_k^s(x)$, $\mathbf{g}_k^t(x)$ and $\mathbf{g}_k^{st}(x)$, respectively.

\subsubsection{Task and Domain Discrimination}
The training objective of task discrimination for $C^s$\&$C^t$ is:
\begin{small}
\begin{equation}
\begin{aligned}
&\min_{C^s, C^t} -\frac{1}{n_s}\sum_{i=1}^{n_s}\log(\mathbf{g}_{y_i^s}^s(x_i^s))-\frac{1}{n_s}\sum_{i=1}^{n_s}\log(\mathbf{g}_{y_i^s}^t(x_i^s))
\end{aligned}
\label{eq1}
\end{equation}
\end{small}The training objective of domain discrimination for $C^{st}$ is:
\begin{small}
\begin{equation}
\min_{C^{st}}-\frac{1}{n_s}\sum_{i=1}^{n_s}\log{\sum_{k=1}^{K}\mathbf{g}_k^{st}(x_i^s)}-\frac{1}{n_t}\sum_{i=1}^{n_t}\log{\sum_{k=1}^{K}\mathbf{g}_{k+K}^{st}(x_i^t)}
\end{equation}
\end{small}

\subsubsection{Category-Level Domain Confusion}
The training objective of category-level confusion is:
\begin{footnotesize}
\begin{equation}
\min_{F}-\frac{1}{2n_s}\sum_{i=1}^{n_s}\log(\mathbf{g}_{y_i^s}^{st}(x_i^s))-\frac{1}{2n_s}\sum_{i=1}^{n_s}\log(\mathbf{g}_{y_i^s+K}^{st}(x_i^s))
\end{equation}
\end{footnotesize}The training objective of domain-level confusion is:
\begin{small}
\begin{equation}
\min_{F}-\frac{1}{2n_t}\sum_{i=1}^{n_t}\log(\sum_{k=1}^{K}\mathbf{g}_k^{st}(x_i^t))-\frac{1}{2n_t}\sum_{i=1}^{n_t}\log(\sum_{k=1}^{K}\mathbf{g}_{k+K}^{st}(x_i^t))
\label{eq4}
\end{equation}
\end{small}Besides, an entropy minimization loss is conducted on $\mathcal{D}_t$ to optimize $F$. For more detailed technical illustration, we recommend referring to the original paper.

\begin{figure*}[t]
\begin{center}
\includegraphics[width=1.\linewidth]{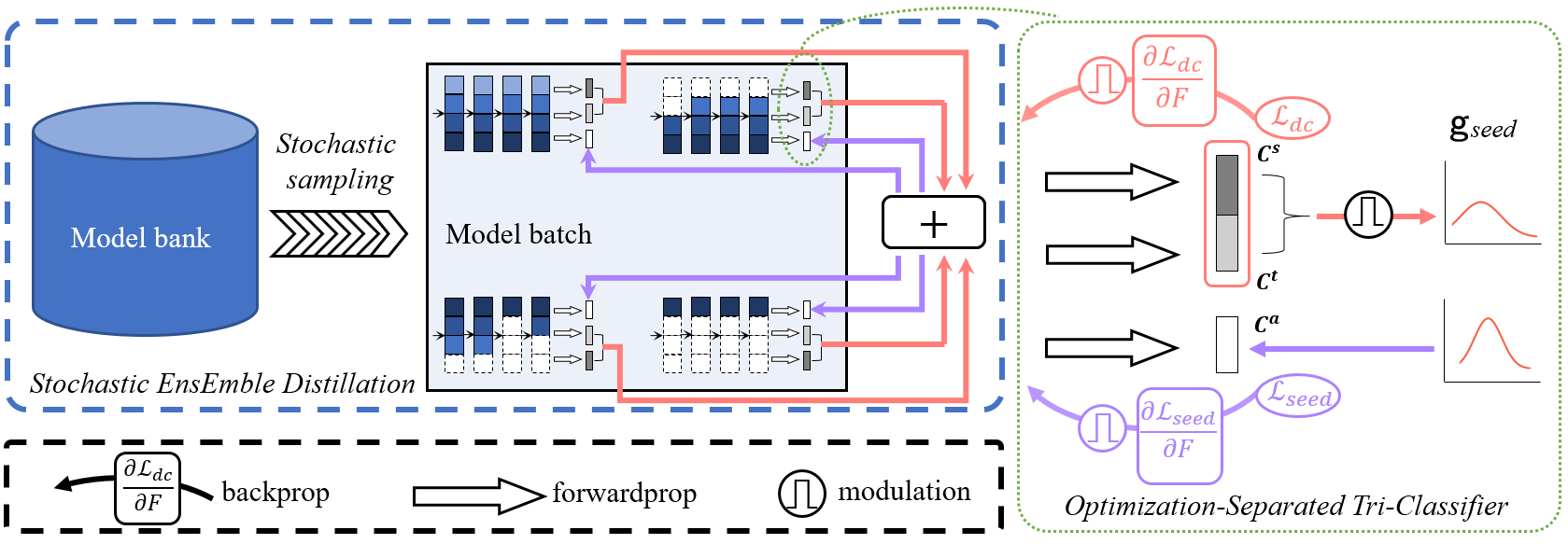}
\end{center}
   \caption{The training details of our proposed SlimDA framework.  Our framework is composed of Stochastic EnsEmble Distillation (SEED) and an Optimization-Separated Tri-Classifier (OSTC) design. The SEED is designed to exploit the complementary knowledge in the model bank for multi-model interactions. The red arrows across $C^s$ and $C^t$ classifiers denote domain confusion training $\mathcal{L}_{dc}$ and knowledge aggregation in the model bank. The purple arrow across $C^a$ classifier denotes SEED optimization $\mathcal{L}_{seed}$.
}
\label{training}
\end{figure*}
\section{Method}

\subsection{Straightforward Baseline}
It has been proven in slimmable neural networks that numerous networks with different widths (i.e., layer channel) can be coupled into a weight-sharing model bank and be optimized simultaneously. 
We begin with a baseline in which SymNet is straightforwardly merged with the slimmable neural networks. The overall objective of SymNet is unified as $\mathcal{L}_{dc}$ for simplicity. 
In each training iteration, several models can be stochastically sampled from the model bank $\{(F_j, C_j^s, C_j^t)\}_{j=1}^m$$\in$$(F, C^s, C^t)$, named as model batch, where $m$ represents the model batch size. Here $(F, C^s, C^t)$ can be viewed as the largest model, and the remaining models can be sampled from it in a weight-sharing manner. To make sure the model bank can be fully trained, the largest and the smallest models\footnote{The smallest model corresponds to 1/64$\times$ FLOPs model (1/8$\times$ channels) by default in this paper.} should be sampled and constituted as a part of model batch in each training iteration. (Note that each model should re-calculate the statistical parameters of BN layers before deploying). 
\begin{footnotesize}
\begin{equation}
\begin{aligned}
&\bigg(\frac{\partial \mathcal{L}_{dc}}{\partial C^s},\frac{\partial \mathcal{L}_{dc}}{\partial C^t}\bigg)\!\!=\!\!\bigg(\frac{1}{m}\sum_{j=1}^{m}\frac{\partial \mathcal{L}_{dc}}{\partial C^s_j}, \frac{1}{m}\sum_{j=1}^{m}\frac{\partial \mathcal{L}_{dc}}{\partial C^t_j}\bigg)
\end{aligned}
\label{1}
\end{equation}
\end{footnotesize}
\begin{footnotesize}
\begin{equation}
\begin{aligned}
&\frac{\partial \mathcal{L}_{dc}}{\partial F}=\frac{1}{m}\sum_{j=1}^{m}\frac{\partial \mathcal{L}_{dc}}{\partial F_j}\\
\end{aligned}
\label{2}
\end{equation}
\end{footnotesize}
This baseline can be viewed as two alternated processes of Eqn.\ref{1} and Eqn.\ref{2} to optimize the model bank. 
To encourage inter-model interaction in the above baseline, we propose our SlimDA framework as shown in Fig. \ref{training}.

\subsection{Stochastic EnsEmble Distillation}
\noindent\textbf{Stochastic Ensemble:~}
It is intuitive that different models in the model bank can learn complementary knowledge about the unlabeled target data. Inspired by Bayesian learning with model perturbation, we exploit the models in the model bank via Monte Carlo Sampling to suppress the uncertainty from unlabeled target data.
The expected prediction $\mathbf{g}_{seed}(x_i^t)$ can be approximated by taking the expectation of $\{\mathbf{g}(x_i;F_j,C_j^s,C_j^t)\}_{j=1}^m$ with respect to the model confidence $\{\mathbf{g}(F_j,C_j^s,C_j^t)\}_{j=1}^m$\footnote{$\mathbf{g}(F_j,C_j^s,C_j^t)$ is short for $\mathbf{g}(F_j,C_j^s,C_j^t\,|\,\mathcal{D})$ where $\mathcal{D}$ denotes the training data. The model confidence, ranging [0,1], can be interpreted to measure the relative accuracy among the models in the model bank.}:
\begin{small}
\begin{equation}
\begin{aligned}
&\mathbf{g}_{seed}(x_i^t)=\mathbb{E}_{\mathbf{g}(F_j,C_j^s,C_j^t)}\big[\mathbf{g}(x_i^t; F_j,C_j^s,C_j^t)\big]
\end{aligned}
\end{equation}
\end{small}
\begin{small}
\begin{equation}
\begin{aligned}
&\mathbf{g}(x_i^t ; F_j,C_j^s,C_j^t)=\frac{1}{2}(\mathbf{g}(x_i^t ; F_j,C_j^s)+\mathbf{g}(x_i^t ; F_j,C_j^t))
\end{aligned}
\end{equation}
\end{small}where $\mathbb{E}$ is a weighted average function with $j$$=$$\{1,...,m\}$, and the subscript of $\mathbb{E}$ denotes the weight. 

\begin{assumption} As demonstrated by extensive empirical results in in-domain generalization work \cite{pmlr-v97-ying19a} and out-of-domain generalization work \cite{2020Learning}, the models with larger capacity\footnote{We use FLOPs as a metric to measure model capacity in this paper.} perform more accurate than those with smaller capacity statistically. Thus, it is reasonable to assume that $\mathbf{g}(F_1,C_1^s,C_1^t)\geq \mathbf{g}(F_2,C_2^s,C_2^t)\geq ... \geq \mathbf{g}(F_m,C_m^s,C_m^t)$ where the index denotes the order of model capacity from large to small. 
\label{assumption}
\end{assumption}
In this work, we empirically define the model confidence in a hard way:
\begin{small}
\begin{equation}
\begin{aligned}
&r_j=\frac{\mathcal{M}(F_j,C_j^s,C_j^t)}{\mathcal{M}(F,C^s,C^t)}\\
&\Omega=\{(F_j,C_j^s,C_j^t) \ \ {\rm whose}\ \  r_j\geq \lambda\}\\
&\mathbf{g}(F_j,C_j^s,C_j^t)=\left\{
\begin{aligned}
&1 \ ,\ {\rm if} \ \ (F_j,C_j^s,C_j^t)\in\Omega  \\
&0 \ ,\ {\rm otherwise}
\end{aligned}
\right.
\end{aligned}
\label{eq:model-confidence}
\end{equation}
\end{small}where $\lambda$ is set 0.5 by default, and $\mathcal{M}(\cdot)$ represents the model capacity. 
As the prediction tends to be uncertain on the unlabeled target data, we aim to produce lower-entropy predictions to boost the discrimination \cite{grandvalet2005semi}. In this work, we apply a sharpening function to $\mathbf{g}_{seed}(x_i^t)$ to induce implicit entropy minimization during SEED training:
\begin{small}
\begin{equation}
\mathbf{g}_{seed,k}(x_i^t)=\mathbf{g}_{seed,k}(x_i^t)^{\frac{1}{\tau}}/\sum_{k'=1}^{K}\mathbf{g}_{seed,k'}(x_i^t)^{\frac{1}{\tau}}
\end{equation}
\end{small}where $\tau$ is a temperature parameter for sharpening, and is set $0.5$ by default in this paper. 
$\mathbf{g}_{seed}(x_i^t)$ is used to refine the model batch, along with domain confusion training in a curriculum mutual learning manner.

\noindent\textbf{Distillation Bridged by Optimization-Separated Tri-Classifier:~} We cannot directly feed $\mathbf{g}_{seed}(x_i^t)$ back to the original bi-classifier for distillation since there exists optimization conflict between intra-model adaptation (Eqn.\ref{eq1}-\ref{eq4}) and inter-model interaction (distillation by $\mathbf{g}_{seed}(x_i^t)$) in the model bank, which are two asynchronous tasks. Specifically,
in the $q$ iteration, domain confusion bi-classifier of multi-models provides two-part information, including task discrimination and domain-confusion, and the two information are aggregated in $\mathbf{g}_{seed}^{q}(x_i^t)$. In the next $q$+1 iteration, the above information can be further updated via the bi-classifier training. However, if we transfer $\mathbf{g}_{seed}^{q}(x_i^t)$ back to the bi-classifier, $\mathbf{g}_{seed}^{q}(x_i^t)$ will offset the gains of the two information in $\mathbf{g}_{seed}^{q+1}(x_i^t)$ and hinder the refinement of $\mathbf{g}_{seed}(x_i^t)$. Thus, the curriculum learning in our SlimDA framework will be destroyed.

To this end, we introduce an Optimization-Separated Tri-Classifier (OSTC) $\{(C_j^s,C_j^t,C_j^a)\}_{j=1}^m$$\in$$(C^s,C^t,C^a)$, where the former two are preserved for domain confusion training, and the last one is designed to receive the stochastically-aggregated knowledge for distillation. The distillation loss is formulated as:
\begin{small}
\begin{equation}
\begin{aligned}
\mathcal{L}_{seed}=&-\frac{1}{m\times n_t}\sum_{j=1}^{m}\sum_{i=1}^{n_t}\mathbf{g}_{seed}(x_i^t)\log(\mathbf{g}(x_i^t;F_j,C_j^a))\\
&-\frac{1}{m\times n_s}\sum_{j=1}^{m}\sum_{i=1}^{n_s}\textbf{1}_{y_i^s}\log(\mathbf{g}(x_i^s;F_j,C_j^a))
\end{aligned}
\end{equation}
\end{small}We optimize $(C^s,C^t,C^a)$ with $\mathcal{L}_{dc}$ and $\mathcal{L}_{seed}$ losses:
\begin{small}
\begin{equation}
\begin{aligned}
\bigg(\frac{\partial \mathcal{L}_{dc}}{\partial C^s},\frac{\partial \mathcal{L}_{dc}}{\partial C^t}\bigg)&=\bigg(\frac{1}{m}\sum_{j=1}^{m}\frac{\partial \mathcal{L}_{dc}}{\partial C^s_j}, \frac{1}{m}\sum_{j=1}^{m}\frac{\partial \mathcal{L}_{dc}}{\partial C^t_j}\bigg)\\
\frac{\partial \mathcal{L}_{seed}}{\partial C^a}&=\frac{1}{m}\sum_{j=1}^{m}\frac{\partial \mathcal{L}_{seed}}{\partial C_j^a}
\label{3}
\end{aligned}
\end{equation}
\end{small}As to optimize $F$, we use the model confidence in Eqn.\ref{eq:model-confidence} to modulate the training objectives of $\mathcal{L}_{dc}$ and $\mathcal{L}_{seed}$:
\begin{small}
\begin{equation}
\begin{aligned}
\begin{split}
\frac{\partial \mathcal{L}_{total}}{\partial F}
=&
\mathbb{E}_{\mathbf{g}(F_j,C_j^s,C_j^t)}\big[\frac{\partial \mathcal{L}_{dc}}{\partial F_j}\big] \\
&+\mathbb{E}_{1-\mathbf{g}(F_j,C_j^s,C_j^t)}\big[\frac{\partial \mathcal{L}_{seed}}{\partial F_j}\big]
\end{split}
\label{4}
\end{aligned}
\end{equation}
\end{small}

To summarize, the OSTC in Eqn.\ref{3} and the feature extractor in Eqn.\ref{4} are optimized in an alternated way in each training iteration. Once finishing training, $(C^s,C^t)$ are discarded and only $C^a$ is retained to deploy more efficiently.

\renewcommand\arraystretch{1} 
\begin{table}[t]
	\centering
	\resizebox{.47\textwidth}{!}{
	\begin{tabular}{c|c|c|c|c|c|c |>{\columncolor{mygray}}c}
	\hline
	Model batch size &I $\to$ P & P $\to$ I & I $\to$ C & C $\to$ I & C $\to$ P & P $\to$ C & Avg.\\
	\hline\hline
	2 (w/ CKD) &71.8&83.3&93.0&82.2&67.3&89.2&81.1\\
	\hline\hline
	2 &73.5&88.3&92.2&87.1&69.3&91.5&83.7\\
	4 & 74.3&89.0&92.6&87.5&69.8&92.0&84.2\\
	6 & 75.3&90.2&94.1&88.3&71.7&94.2&85.6\\
	8 &76.1 &89.9&94.9&88.1&71.7&94.2&85.8\\
	10 & 78.3&90.7&95.8&88.3&71.8&94.8&86.6\\
	\hline
	\end{tabular}
	}
	\vspace{-3mm}
	\caption{Two ablation studies on ImageCLEF-DA: 1) The second row represents the results from conventional knowledge distillation (CKD). 2) Comparison among different model batch sizes in SlimDA. We report the above results for $1/64\times$ models.}
	\vspace{-2mm}
	\label{modelbatchsize}
\end{table}
\begin{table}[t]
    \large
	\centering
	\resizebox{.48\textwidth}{!}{
	\begin{tabular}{ccc|c|c|c|c|c|c|c}
	\hline
	Baseline&SEED & OSTC & 1$\times$ &  1/2$\times$ & 1/4$\times$ & 1/8$\times$ & 1/16$\times$ & 1/32$\times$ & 1/64$\times$ \\
	\hline\hline
	\checkmark&& &88.6&88.0&86.9&86.0&84.1&82.0&81.7 \\
	\checkmark&\checkmark& &87.9&87.6&87.3&86.9&86.5&86.1&85.9 \\
	\checkmark&\checkmark&\checkmark& 88.9&88.7&88.8&88.4&88.3&87.2&86.6 \\
	\hline
	\end{tabular}
	}
	\vspace{-3mm}
	\caption{The ablation studies for the components in SlimDA on ImageCLEF-DA dataset.}
	\label{components:imagecled-da}
	\vspace{-2mm}
\end{table}
\renewcommand\arraystretch{1} 
\begin{table}[t]
	\centering
	\resizebox{.48\textwidth}{!}{
	\begin{tabular}{l|c|c|c|c|c|c|c}
	\hline
	Methods & 1$\times$ &  1/2$\times$ & 1/4$\times$ & 1/8$\times$ & 1/16$\times$ & 1/32$\times$ & 1/64$\times$ \\
	\hline\hline
	SymNet w/o SlimDA           &78.9&77.0 &76.7 &74.8 & 71.2& 68.2&69.3  \\
	
	SymNet w/ SlimDA &79.2 &79.0 &79.0 &78.7 &78.8 &78.2&78.3  \\
	\hline
	\textbf{improvement}&\textbf{0.3$\uparrow$}&\textbf{2.0$\uparrow$}&\textbf{2.3$\uparrow$}&\textbf{3.9$\uparrow$}&\textbf{7.6$\uparrow$}&\textbf{10.0$\uparrow$}&\textbf{9.0$\uparrow$}\\
	\hline\hline
	MCD w/o SlimDA           &77.2&75.0&75.0&72.3&70.3&68.7&69.6  \\
	
	MCD w/ SlimDA &78.5&78.2&78.1&78.0&77.7&77.6&77.7  \\
	\hline
	\textbf{improvement} &\textbf{1.3$\uparrow$}&\textbf{3.2$\uparrow$}&\textbf{3.1$\uparrow$}&\textbf{5.7$\uparrow$}&\textbf{7.4$\uparrow$}&\textbf{8.9$\uparrow$}&\textbf{8.1$\uparrow$}  \\
	\hline\hline
	STAR w/o SlimDA           &76.9&74.0&69.7&68.1&65.6&62.9&64.7  \\
	
	STAR w/ SlimDA &77.8&77.5&77.2&77.2&77.0&76.9&77.0  \\
	\hline
	\textbf{improvement}&\textbf{0.9$\uparrow$}&\textbf{3.5$\uparrow$}&\textbf{7.5$\uparrow$}&\textbf{9.1$\uparrow$}&\textbf{11.4$\uparrow$}&\textbf{14.0$\uparrow$}&\textbf{12.3$\uparrow$}  \\
	\hline
	\end{tabular}
	}
	\vspace{-3mm}
	\caption{Two ablation studies on ImageCLEF-DA for I$\to$P adaptation task: 1) Comparison with different UDA methods injected into SlimDA, which indicates the universality of our framework. 2) Comparison with stand-alone networks (aka ``w/o SlimDA''), which have the same architectures as our searched models but are trained individually outsides the model bank.}
	\label{standalone}
	\vspace{-2mm}
\end{table}
\begin{table}[t]
\Huge
\begin{center}
\resizebox{0.48\textwidth}{!}{
\begin{tabular}{l|c|c|c|c|c|c|c|c |>{\columncolor{mygray}} c}
\toprule
Methods &\#Params & FLOPs & I $\to$ P & P $\to$ I&I $\to$ C&C $\to$ I&C $\to$ P&P $\to$ C &Avg. \\ 
\hline\hline
ShuffleNetV2 \cite{ma2018shufflenet}& 2.3M&146M &63.2&65.4&82.1&81.0&65.2&78.9&72.6 \\
MobileNetV3 \cite{howard2019searching}&5.4M &219M &72.8&85.3&93.2&80.3&65.0&91.7&81.4 \\
GhostNet \cite{han2020ghostnet}&5.2M &141M &75.8&89.5&95.5&86.1&70.2&94.0&85.2 \\

MobileNetV2 \cite{sandler2018mobilenetv2} &3.5M &300M &76.0&90.6&95.1&87.0&69.1&95.1&85.5 \\
EfficientNet\_B0 \cite{tan2019efficientnet} &5.3M&390M &76.5&88.5&96.5&87.3&71.3&94.0&85.6 \\
\hline\hline
SlimDA ($1/8\times$ResNet-50)&4.0M &517M &78.7 &91.7&97.2&90.5&75.8&96.2&88.4 \\
SlimDA ($1/64\times$ResNet-50)&1.6M&64M &78.3 &90.7&95.8&88.3&71.8&94.8&86.6 \\
\bottomrule
\end{tabular}
}
\end{center}
\vspace{-6mm}
\caption{Performance comparison with different state-of-the-art lightweight networks on ImageCLEF-DA dataset.}
\vspace{-3mm}
\label{lighnet:imageclef}
\end{table}
\vspace{-0.8mm}

\subsection{Unsupervised Performance Evaluation Metric}
In the context of UDA, one of the challenging points is to evaluate the performance ranking of models on the unlabeled target data rather than the searching methods. 
According to the \emph{Triangle Inequality Theorem}, we can obtain the relationship among the predictions of the candidate model $(F_j,C_j^a)$, the largest model $(F,C^a)$, as well as the ground-truth label:
\begin{small}
\begin{equation}
\begin{aligned}
\begin{split}
&\parallel \mathbf{g}(\mathcal{D}_t;F_j,C_j^a)\!\!-\!\!GT(\mathcal{D}_t)\parallel_2^2 < \parallel \mathbf{g}(\mathcal{D}_t;F,C^a)\!\!-\!\!GT(\mathcal{D}_t)\parallel_2^2 \\
&\!\!+\!\!\parallel \mathbf{g}(\mathcal{D}_t;F_j,C_j^a)\!\!-\mathbf{g}(\mathcal{D}_t;F,C^a)\parallel_2^2
\end{split}
\label{6}
\end{aligned}
\end{equation}
\end{small}where $GT(\mathcal{D}_t)$ denotes the ground-truth label of the target data. 
According to the assumption \ref{assumption}, the model with the largest capacity tends to be the most accurate in the model bank, which means:
\begin{small}
\begin{equation}
\parallel \mathbf{g}(\mathcal{D}_t;F_j,C_j^a)\!\!-\!\!GT(\mathcal{D}_t)\parallel_2^2 >\parallel \mathbf{g}(\mathcal{D}_t;F,C^a)\!\!-\!\!GT(\mathcal{D}_t)\parallel_2^2
\label{7}
\end{equation}
\end{small}Combining Eqn.\ref{6} and Eqn.\ref{7}, we can take the model with the largest capacity as an anchor to compare the performance of candidate models on the unlabeled target data. 
The Unsupervised Performance Evaluation Metric (UPEM) for each model can be written as:
\begin{small}
\begin{equation}
\Delta_j=\parallel \mathbf{g}(\mathcal{D}^t;F_j,C_j^a)-\mathbf{g}(\mathcal{D}^t;F,C^a)\parallel_2^2
\vspace{-1mm}
\end{equation}
\end{small}where $\Delta_j$ is the $\mathcal{L}$2 distance between outputs of the candidate model and the anchor model.
With the UPEM, we can use a greedy search method \cite{Yu2019NetworkSB,2020Neural} for neural architecture search (Note that we can also use other search methods, but this is not the deciding point in this paper).

\begin{table*}[t]
\tiny
\begin{center}
\resizebox{.97\textwidth}{!}{
\begin{tabular}{l|r|r|c|c|c|c|c|c |>{\columncolor{mygray}} c|>{\columncolor{mygray}}c}
\hline
Methods &\#Params & FLOPs & I $\to$ P & P $\to$ I & I $\to$ C & C $\to$ I & C $\to$ P & P $\to$ C & Avg.& $\Delta$ \\ 
\hline\hline
Source Only \cite{SymNet} & 1$\times$ & 1$\times$ &74.8 & 83.9 & 91.5 & 78.0 & 65.5 & 91.2 & 80.7&--\\
DAN \cite{MMD} & 1$\times$ & 1$\times$ &74.5 & 82.2 & 92.8 & 86.3 & 69.2 & 89.8 & 82.5&--\\
RevGrad \cite{GRL} & 1$\times$ & 1$\times$ &75.0 & 86.0 & 96.2 & 87.0 & 74.3 & 91.5 & 85.0&--\\
MCD \cite{MCD} (impl.)&1$\times$ & 1$\times$  &77.2&87.2&93.8&87.7&71.8&92.5&85.0&--\\
STAR \cite{STAR} (impl.)&1$\times$ & 1$\times$  &76.9&87.7&93.8&87.6&72.1&92.7&85.1&--\\
CDAN+E \cite{Long2017Conditional} & 1$\times$ & 1$\times$ &77.7 & 90.7 & 97.7 & 91.3 & 74.2 & 94.3 & 87.7&--\\
SymNets \cite{SymNet} & 1$\times$ & 1$\times$ &80.2 & 93.6 & 97.0 & 93.4 & 78.7 & 96.4 & 89.9&--\\
SymNets (impl.) & 1$\times$ & 1$\times$ &78.8 & 92.2 & 96.7 & 91.0 & 76.0 & 96.2 & 88.5&--\\
\hline\hline
TCP \cite{Yu2019AcceleratingDU} & 1/1.7$\times$ & \makecell[c]{--} &75.0 & 82.6 &     92.5    &   80.8 & 66.2  &  86.5 & 80.6 &--\\
 & 1/2.5$\times$ & \makecell[c]{--} & 67.8  &  77.5 & 88.6 & 71.6 & 57.7  &79.5&73.8  &--\\
\hline
ADMP \cite{Feng2020ADMPAA} &1/1.7$\times$& \makecell[c]{--} & 77.3 & 90.2 &  90.2  &95.8 & 88.9 & 73.7  &  86.3   &--\\
&1/2.5$\times$& \makecell[c]{--}&77.0 & 89.5 & 95.5 &  88.9   &72.3  & 91.2 & 85.7 &--\\
\hline\hline
SlimDA&1$\times$&1$\times$&79.2 &92.3&97.5&91.2&76.7&96.5&88.9&--\\
&1/1.9$\times$&1/2$\times$  &79.0 &92.3&97.3&90.8&76.8&96.2&88.7&0.2$\downarrow$\\
&1/3.9$\times$&1/4$\times$  &79.0 &92.2&97.3&90.8&77.2&96.3&88.8&0.1$\downarrow$\\
&1/9.4$\times$&1/8$\times$  &78.7 &91.7&97.2&90.5&75.8&96.2&88.4&0.5$\downarrow$\\
&1/12.8$\times$&1/16$\times$  &78.8 &91.5&97.3&90.2&76.0&96.2&88.3&0.6$\downarrow$\\
&1/28.8$\times$&1/32$\times$ &78.2 &90.5&96.7&89.3&72.2&96.0&87.2&1.7$\downarrow$\\
&1/64$\times$&1/64$\times$ &78.3 &90.7&95.8&88.3&71.8&94.8&86.6&2.3$\downarrow$\\
\hline
\end{tabular}
}
\end{center}
\vspace{-6mm}
\caption{Performance on the ImageCLEF-DA dataset. ``--'' means that the results are not reported in the original paper. ``impl.'' denotes our re-implementation using the released code. ``$\Delta$'' indicates the performance gap between the searched model and ResNet-50 based model for each UDA method. TCP and ADMP are two related cross-domain network compression methods. We adapt the architectures for six adaptation tasks under seven computational constraints (FLOPs). Since the model architectures adapted for different tasks are different even with the same FLOPs, we calculate the parameter reduction (\#Params) by averaging models in 6 adaptation tasks.}
\vspace{-2mm}
\label{clef}
\end{table*}

\begin{table*}[t]
\tiny
\begin{center}
\resizebox{1.0\textwidth}{!}{
\begin{tabular}{l|r|r|c|c|c|c|c|c| >{\columncolor{mygray}} c|>{\columncolor{mygray}}c}
\hline
Methods &\#Params &FLOPs &A $\to$ W & D$\to$W & W$\to$D & A$\to$D & D$\to$A & W$\to$A & Avg.&$\Delta$ \\ 
\hline\hline
Source Only \cite{SymNet} & $1\times$ & $1\times$ & 79.9 & 96.8 & 99.5 & 84.1 & 64.5 & 66.4 & 81.9&--\\
Domain Confusion \cite{Hoffman2015SimultaneousDT}&$1\times$ & $1\times$& 83.0 & 98.5 & 99.8 & 83.9 & 66.9 & 66.4 & 83.1&--\\
Domain Confusion+Em \cite{Hoffman2015SimultaneousDT}&$1\times$ & $1\times$& 89.8 & 99.0 & 100.0 & 90.1 & 73.9 & 69.0 & 87.0&--\\
BNM \cite{cui2020towards} &$1\times$ & $1\times$&91.5&98.9&100.0&90.3&70.9&71.6&87.1&--\\
DMP \cite{Luo2020UnsupervisedDA}&$1\times$ & $1\times$&93.0&99.0&100.0&91.0&71.4&70.2&87.4&--\\
DMRL \cite{Wu2020DualMR}&$1\times$ & $1\times$&90.8&99.0&100.0&93.4&73.0&71.2&87.9&--\\

SymNets \cite{SymNet} &$1\times$ & $1\times$& 90.8 & 98.8 & 100.0 & 93.9 & 74.6 & 72.5 & 88.4 &--\\
SymNets (impl.) &$1\times$ & $1\times$& 91.0 & 98.4 & 99.6 & 89.7 & 72.2 & 72.5 & 87.2 &--\\
\hline\hline
TCP \cite{Yu2019AcceleratingDU} & 1/1.7$\times$ & \makecell[c]{--} & 81.8& 98.2  &99.8 &  77.9  & 50.0 & 55.5  & 77.2 &--\\
&  1/2.5$\times$ &  \makecell[c]{--} & 77.4 &96.3 &100.0 &   72.0  &  36.1  & 46.3&  71.3&--\\
\hline
ADMP \cite{Feng2020ADMPAA} &1/1.7$\times$ & \makecell[c]{--}& 83.3  &   98.9 & 99.9 & 83.1 &63.2 &64.2 &82.0&--\\
& 1/2.5$\times$ &\makecell[c]{--} & 82.1  &  98.6  &  99.9 &  81.5   &  63.0 & 63.2   &  81.3    &--\\
\hline\hline
SlimDA& 1$\times$ &1$\times$   &90.7&91.2&91.1&99.8&73.7&71.0&87.6&--\\
&1/2$\times$&1/2$\times$ &90.7 &99.1&100.0&91.8&73.3&71.1&87.6&0.0$\downarrow$\\
&1/4$\times$&1/4$\times$ &90.5 &98.1&99.8&91.9&73.1&71.2&87.4&0.2$\downarrow$\\
&1/10$\times$&1/8$\times$ & 90.6&98.8&100.0&91.6&72.9&71.1&87.5&0.1$\downarrow$\\
&1/14$\times$&1/16$\times$&90.5 &98.7&99.8&91.4&73.1&71.0&87.4&0.2$\downarrow$\\
&1/20$\times$&1/32$\times$&90.8 &97.7&99.5&91.5&71.8&70.8&87.0&0.6$\downarrow$\\
&1/64$\times$&1/64$\times$&91.2 &97.2&98.9&91.2&71.3&68.9&86.8&0.8$\downarrow$\\
\hline
\end{tabular}
}
\end{center}
\vspace{-6mm}
\caption{Performance on the Office-31 dataset. As for other illustrations, please refer to the caption of Table \ref{clef}.}
\vspace{-4mm}
\label{office31}
\end{table*}

\section{Experiments}
\subsection{Dataset}

\noindent\textbf{ImageCLEF-DA} \cite{MMD} consists of 1,800 images with 12 categories over three domains: Caltech-256 (C), ImageNet ILSVRC 2012 (I), and Pascal VOC 2012 (P).

\noindent\textbf{Office-31} \cite{saenko2010adapting} is a popular benchmark with about 4,110 images sharing 31 categories of daily objects from 3 domains: Amazon (A), Webcam (W) and DSLR (D).

\noindent\textbf{Office-Home} \cite{venkateswara2017deep} contains 15,500 images sharing 65 categories of daily objects from 4 different domains: Art (Ar), Clipart (Cl), Product (Pr), and Real-World (Rw).

\subsection{Model Bank Configurations}
\vspace{-2mm}
Following the exiting methods \cite{ijcai2019-291,Feng2020ADMPAA,Yu2019AcceleratingDU}, we select ResNet-50 \cite{resnet} as the main network to conduct the following experiments. Unlike these methods, the ResNet-50 adopted in this paper is a super-network that couples numerous models with different layer widths to form the model bank. Identical to these methods, the super-network should be firstly pre-trained on ImageNet and then fine-tuned on the downstream tasks.

\subsection{Implementation Details}
An SGD optimizer with momentum of 0.9 is adopted to train all UDA tasks in this paper. Following \cite{SymNet}, the learning rate is adjusted by $l=l_0/(1+\alpha p)^\beta$, where $l_0=0.01$, $\alpha=10$, $\beta=0.75$, and $p$ varies from $0$ to $1$ linearly with the training epochs. The training epoch is set 40. The training and testing image resolution is $224\times 224$. An important technical detail is that, before performance evaluation, the models sampled in the model bank should update the statistic of their BN layers on the target domain via AdaBN \cite{AdaBN}. We mainly take the computational complexity (FLOPs) as resource constraint for architecture adaptation, and set 1/64$\times$ FLOPs as the smallest model by default.

\begin{figure}[t]
\vskip -0.0in
\begin{center}
\centerline{\includegraphics[width=1.0\columnwidth]{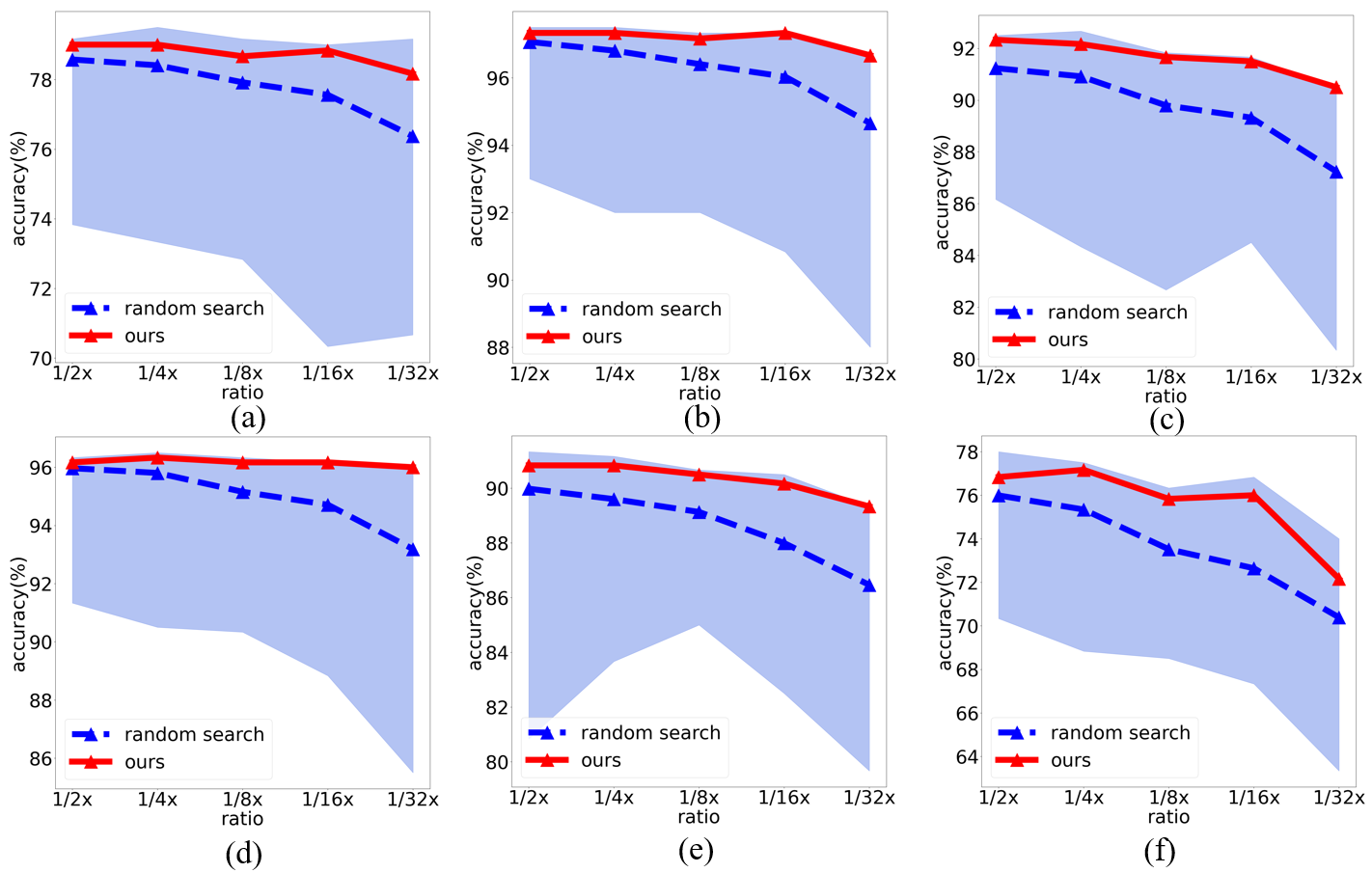}}
\vspace{-3mm}
\caption{Comparison with randomly-searched models on six adaptation tasks on ImageCLEF-DA. One hundred models are randomly-searched under each FLOPs. The blue-filled area represents the gap between the maximum and minimum accuracy among random-searched models. The dotted blue dotted line represents the averaged values of random-searched models, and the solid red line represents the accuracies of our searched models.}
\label{randomsearch}
\end{center}
\vspace{-11.0mm}
\end{figure}

\begin{table*}[t]
\begin{center}
\resizebox{1.0\textwidth}{!}{
\begin{tabular}{l|r|c|c|c|c|c|c|c|c|c|c|c|c| >{\columncolor{mygray}} c|>{\columncolor{mygray}} c}
\hline
Methods & FLOPs & Ar$\to$Cl & Ar$\to$Pr & Ar$\to$ Rw & Cl$\to$Ar & Cl$\to$Pr & Cl$\to$Rw & Pr$\to$Ar & Pr$\to$Cl & Pr$\to$Rw & Rw$\to$Ar & Rw$\to$Cl & Rw$\to$Pr & Avg. &$\Delta$ \\
\hline\hline
Source Only & $1\times$ & 34.9 & 50.0 & 58.0 & 37.4 & 41.9 & 46.2 & 38.5 & 31.2 & 60.4 & 53.9 & 41.2 & 59.9 & 46.1&--\\
DAN & $1\times$  & 43.6& 57.0 &67.9& 45.8 &56.5& 60.4 &44.0& 43.6 &67.7 &63.1&51.5 &74.3& 56.3&--\\
RevGrad  & $1\times$ & 45.6 &59.3& 70.1& 47.0 &58.5 &60.9 &46.1& 43.7& 68.5& 63.2 &51.8& 76.8& 57.6&--\\
CDAN-E  & $1\times$ & 50.7 &70.6 &76.0 &57.6 &70.0 &70.0 &57.4 &50.9 &77.3 &70.9 &56.7 &81.6 &65.8&--\\
SymNet & $1\times$ & 47.7 &72.9 &78.5 &64.2 &71.3 &74.2 &64.2 &48.8 &79.5 &74.5 &52.6 &82.7& 67.6 &--\\
BNM & $1\times$ &52.3&73.9&80.0&63.3&72.9&74.9&61.7&49.5&79.7&70.5&53.6&82.2&67.9 &--\\
\hline\hline
SlimDA &  1$\times$ &52.8 &72.3&77.2&63.5&72.3&73.3&64.8&52.2&79.7&72.4&57.8&82.8&68.4&--\\
&  1/2$\times$  &52.4 &72.0&77.2&62.8&72.0&73.0&65.1&53.2&79.4&72.1&55.3&82.0&68.0&0.4$\downarrow$\\
&  1/4$\times$  &51.9 &71.8&77.1&62.4&71.9&72.3&64.8&53.1&78.8&71.9&55.1&82.1&67.8&0.6$\downarrow$\\
&  1/8$\times$  &51.6 &71.4&76.6&62.5&71.0&71.0&65.0&53.0&78.4&71.5&55.2&81.6&67.4&1.0$\downarrow$\\
&  1/16$\times$ &51.0 &71.0&75.9&61.3&70.6&70.0&64.4&52.3&77.7&70.2&54.6&81.2&66.7&1.7$\downarrow$\\
&  1/32$\times$ &50.0 &70.6&74.0&57.7&70.3&68.9&60.1&51.6&76.3&67.5&51.7&80.9&65.0&3.4$\downarrow$\\
&  1/64$\times$ &49.7 &70.1&72.9&56.6&70.0&66.3&56.5&48.3&75.9&65.9&55.5&80.9&64.0&4.4$\downarrow$\\
\hline
\end{tabular}
}
\end{center}
\vspace{-6mm}
\caption{Performance on the Office-Home dataset. As for other illustrations, please refer to the caption of Table \ref{clef}.}
\label{officehome}
\vspace{-3mm}
\end{table*}

\subsection{Ablation Studies}
\subsubsection{Analysis for SEED}
\noindent\textbf{Comparison with Conventional Knowledge Distillation:~}As shown in Table \ref{modelbatchsize}, we can observe that our SEED with different model batch size can outperform the conventional knowledge distillation by a large margin even under 1/64$\times$ FLOPs of ResNet-50 on ImageCLEF-DA.

\noindent\textbf{Comparison among different model batch sizes:~} Model batch size is a vital hyper-parameter of our framework. Intuitively, a larger model batch is more sufficient to approximate the knowledge aggregation in the model bank. As shown in Table \ref{modelbatchsize}, a large model batch size is beneficial for our optimization method. We set the model batch size as 10 by default in our following experiments.

\noindent\textbf{Comparison with stand-alone training:~} 
As shown in Table \ref{standalone}, compared with stand-alone training with the same network configuration, SEED can improve the overall adaptation performance by a large margin. Here ``stand-alone'' means that the models from $1\times$ to $1/64\times$ with the same topological configuration to the SlimDA counterparts are adapted individually outside the model bank. Specifically, not only the tiny models (from $1/2\times$ to $1/64\times$), but also the $1\times$ model trained with SEED outperform the corresponding stand-alone ones, which can be supported by more results in Table \ref{clef} and Table \ref{office31} (comparing the performance of $1\times$ models and our re-implemented ones).

\noindent\textbf{Effectiveness of each component in our SlimDA:~} We conduct ablation studies to investigate the effectiveness of the components in our SlimDA framework. As shown in Table \ref{components:imagecled-da}, the second row ``Baseline'' denotes the approach merging SymNet and slimmable neural network straightforwardly. We can observe that the SEED has a significant impact on the performance with fewer FLOPs, such as 1/4$\times$, 1/8$\times$, 1/16$\times$, 1/32$\times$, and 1/64$\times$, but the performances of 1$\times$ and 1/2 $\times$ models with SEED 
fall 0.7\% and 0.4\%, respectively, compared with the baseline, which is attributed to the optimization conflict between intra-model domain confusion and inter-model SEED. The last row shows that our proposed OSTC provides an impressive improvement on the performance of large models (1$\times$ and 1/2$\times$) compared with both the SEED and baseline. Moreover, our proposed OSTC can further improve the performance of other models with fewer FLOPs.
Overall, each component in SlimDA contributes to the performance-boosting of tiny models (from 1/64$\times$ to 1/4$\times$), the SEED w/o OSTC indeed brings negative transferring for models with larger capacity. However, our proposed OSTC can remedy the negative transferring issue and provide additional boosting under six FLOPs settings. The results in the ablation study also demonstrate the process of solving the challenges in our framework to accomplish the combination between UDA training and weight-sharing model bank.

\vspace{-6mm}
\subsubsection{Analysis for Architecture Adaptation} 

\begin{figure}[t]
\vskip -0.0in
\begin{center}
\centerline{\includegraphics[width=1.0\columnwidth]{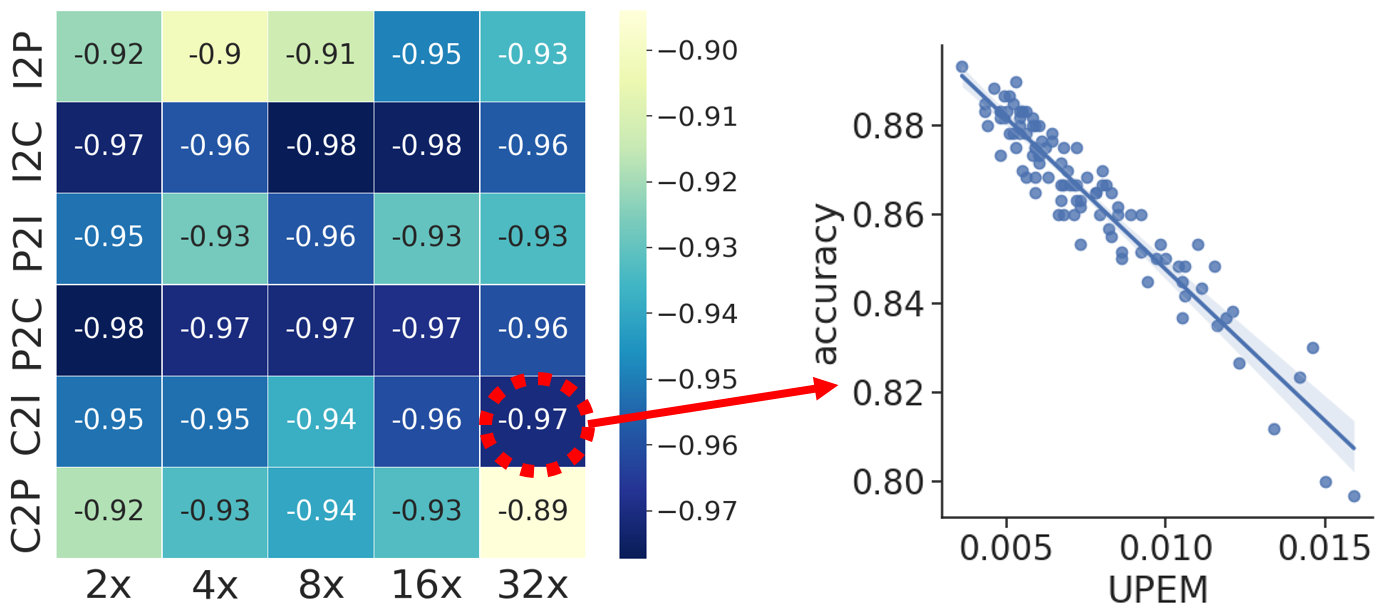}}
\vspace{-2mm}
\caption{Pearson correlation coefficient between the unsupervised performance evaluation metric (UPEM) and the accuracy using ground-truth labels. It is performed on six adaptation tasks (ImageCLEF-DA) under five different computation constraints. In each grid, we sample 100 models to calculate the Pearson correlation coefficient. If it gets close to $-1$, it means that our metric is identical to use the ground-truth labels to measure the performance of each model.}
\vspace{-4mm}
\label{metric}
\end{center}
\vspace{-7mm}
\end{figure}

\noindent\textbf{Analysis for UPEM:~}
To validate the effectiveness of our proposed UPEM, we conduct 30 experiments which are composed of six adaptation tasks and five computational constraints. In each experiment, we sample 100 models in the model bank to evaluate the correlation between UPEM and the supervised accuracy with the ground-truth labels on target data. As shown in Fig.\ref{metric}, the Pearson correlation coefficients almost get close to $-1$, which means the smaller the UPEM is, the higher the accuracy is.

\noindent\textbf{Comparison with Random Search:} As shown in Fig.\ref{randomsearch}, the necessity of architecture adaptation is highlighted
from the significant performance gap among random searched models under each computational budget.
Meanwhile, the architectures searched for different  tasks are different. Moreover, given the same computational budgets, models greedily searched from the model bank with our UPEM are the top series in their actual performance sorting.

\vspace{-1mm}
\subsubsection{Compatibility with Different UDA Methods} 
Our SlimDA can also be cooperated with other UDA approaches, such as MCD \cite{MCD} and STAR \cite{STAR}. To present the universality of our framework, we carry out additional experiments of SlimDA with MCD and STAR, as well as their stand-alone counterparts. As shown in Table \ref{standalone}, the results from our SlimDA with MCD and STAR are still superior to the corresponding stand-alone counterparts by a large margin, which further validates the effectiveness of our SlimDA framework from another perspective.

\vspace{-2mm}
\subsubsection{Convergence Performance} 
\begin{figure}[t]
\begin{center}
   \includegraphics[width=1.0\linewidth]{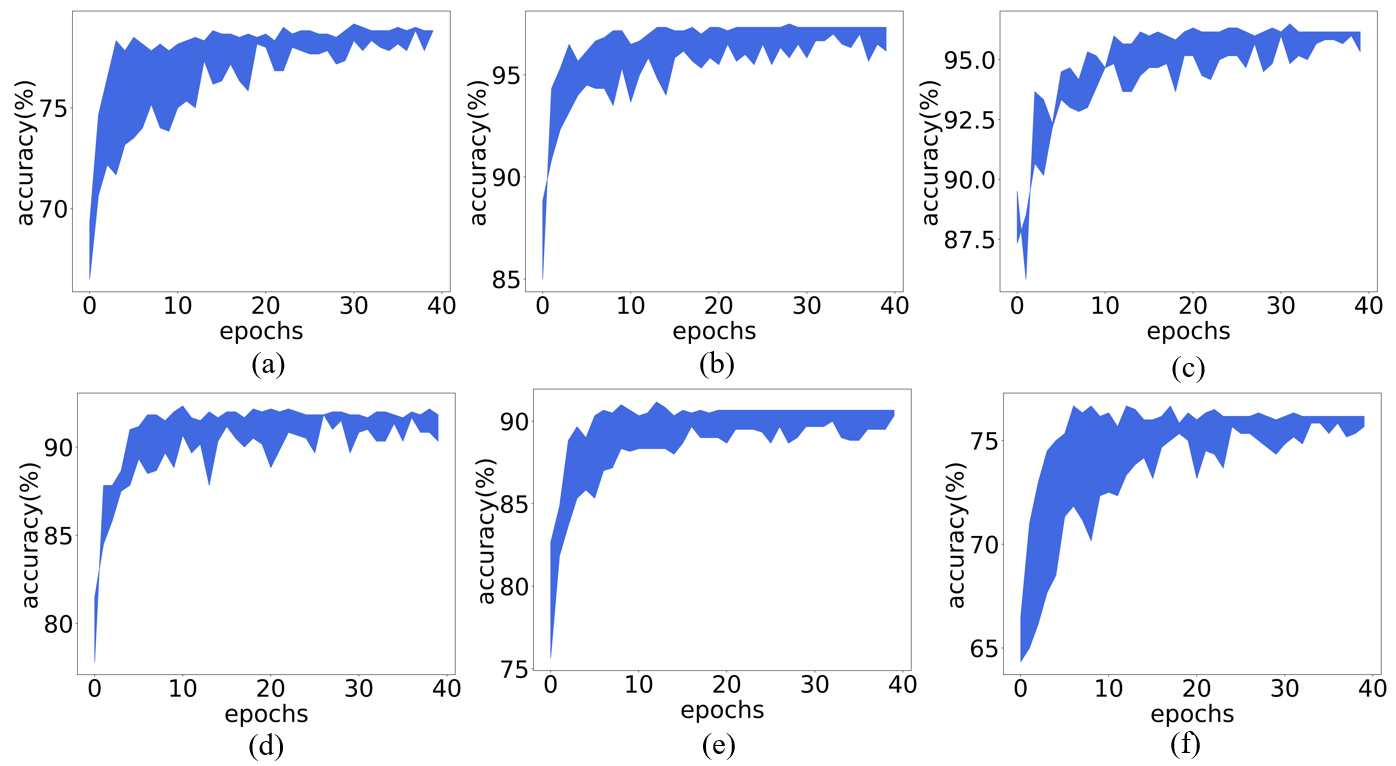}
\end{center}
\vspace{-5mm}
   \caption{Convergence performance of the model bank. Six sub-figures correspond to six adaptation tasks on ImageCLEF-DA. The results in each epoch are from 10 randomly-sampled models.}
\vspace{-3mm}
\label{Convergence}
\end{figure}

We sample different models for performance evaluation during model bank training. As shown in Fig.\ref{Convergence}, the models coupled in the model bank are optimized simultaneously. The performances are steadily improved as learning goes.

\subsection{Comparisons with State-of-The-Arts}

\noindent\textbf{Comparison with UDA methods:~}
We conduct extensive experiments on three popular domain adaptation datasets, including ImageCLEF-DA, Office-31, and Office-Home. Detailed results and performance comparison with SOTA UDA methods are presented in Table \ref{clef}, Table \ref{office31} and Table \ref{officehome}. The benefits of our proposed SlimDA can be further pronounced when we reduce FLOPs extremely. Our method can preserve the accuracy improvement against the source-only model even when reducing FLOPs up to 1/64$\times$. As we can see in Table \ref{clef}, Table \ref{office31} and Table \ref{officehome}, when we reduce FLOPs even up to 1/64$\times$, the accuracy on ImageCLEF-DA, Office-31 and Office-Home datasets only drop about 2.3\%, 0.8\% and 4.4\%, respectively.

\noindent\textbf{Comparision with cross-domain network compression methods:~}
TCP \cite{Yu2019AcceleratingDU} and ADMP \cite{Feng2020ADMPAA} are two main cross-domain network compression methods for channel numbers, our method can significantly outperform them by a very large margin and achieve new SOTA results. 

\noindent\textbf{Comparison with lightweight networks:~}
As shown in Table \ref{lighnet:imageclef}, not only human-designed lightweight networks like MobileNet series \cite{sandler2018mobilenetv2,howard2019searching}, ShuffleNet \cite{ma2018shufflenet} and GhostNet \cite{han2020ghostnet}, but also auto-designed architectures like EfficientNet \cite{tan2019efficientnet} is over-passed by our SlimDA with fewer FLOPs/parameters.

\section{Conclusions}
In this paper, we propose a simple yet effective SlimDA framework to facilitate weight and architecture joint-adaptation.
In SlimDA, the proposed SEED exploits architecture diversity in a weight-sharing model bank to suppress prediction uncertainty on the unlabeled target data, and the proposed OSTC modulates the optimization conflict between intra-model adaptation and inter-model interaction. In this way, we can flexibly distribute resource-satisfactory models via a retrain-free sampling manner to various devices on target domain. Moreover, we propose UPEM to select the optimal cross-domain model under each computational budget.
Extensive ablation studies and experiments are carried out to validate the effectiveness of SlimDA. Our SlimDA can also be extended to other visual computing tasks, such as object detection and semantic segmentation, which we leave as our future work. 
In short, our work provides a practical UDA framework towards real-world scenario, and we hope it can bring new inspirations to the widespread application of UDA.

\vspace{2mm}
\noindent\textbf{Limitations.} The assumption \ref{assumption} is an essential prerequisite in this work, which will hold at least to a reasonable extent. However, it is not guaranteed theoretically, which may invalidate our method in some unknown circumstances.

\section*{Acknowledgements} 
This work was sponsored in part by National Natural Science Foundation of China (62106220, U20B2066),  Hikvision Open Fund,
AI Singapore (Award No.: AISG2-100E-2021-077), and MOE AcRF TIER-1 FRC RESEARCH GRANT (WBS: A-0009456-00-00).

\clearpage
%%%%%%%%% REFERENCES
{\small
\bibliographystyle{ieee_fullname}
\bibliography{egbib}
}

\appendix

\section{Implementation Details}

\subsection{Details for SEED}
The overall SEED schema is summarized in Algorithm \ref{algo:seed}. For each data batch, we sample a model batch stochastically and then utilize SEED to train each model batch. We first calculate model confidence in each model batch, which has two-fold utilities in the SEED: 1) weighting the predictions from different models to generate more confident ensemble prediction; 2) weighting the gradients from intra-model adaptation and inter-model interaction losses with respect to the feature extractors.
After SEED training, we can obtain a ``once-for-all'' domain adaptive model bank.

\subsection{Details for Neural Architecture Search}
\begin{figure}[t]
\begin{center}
\includegraphics[width=.95\linewidth]{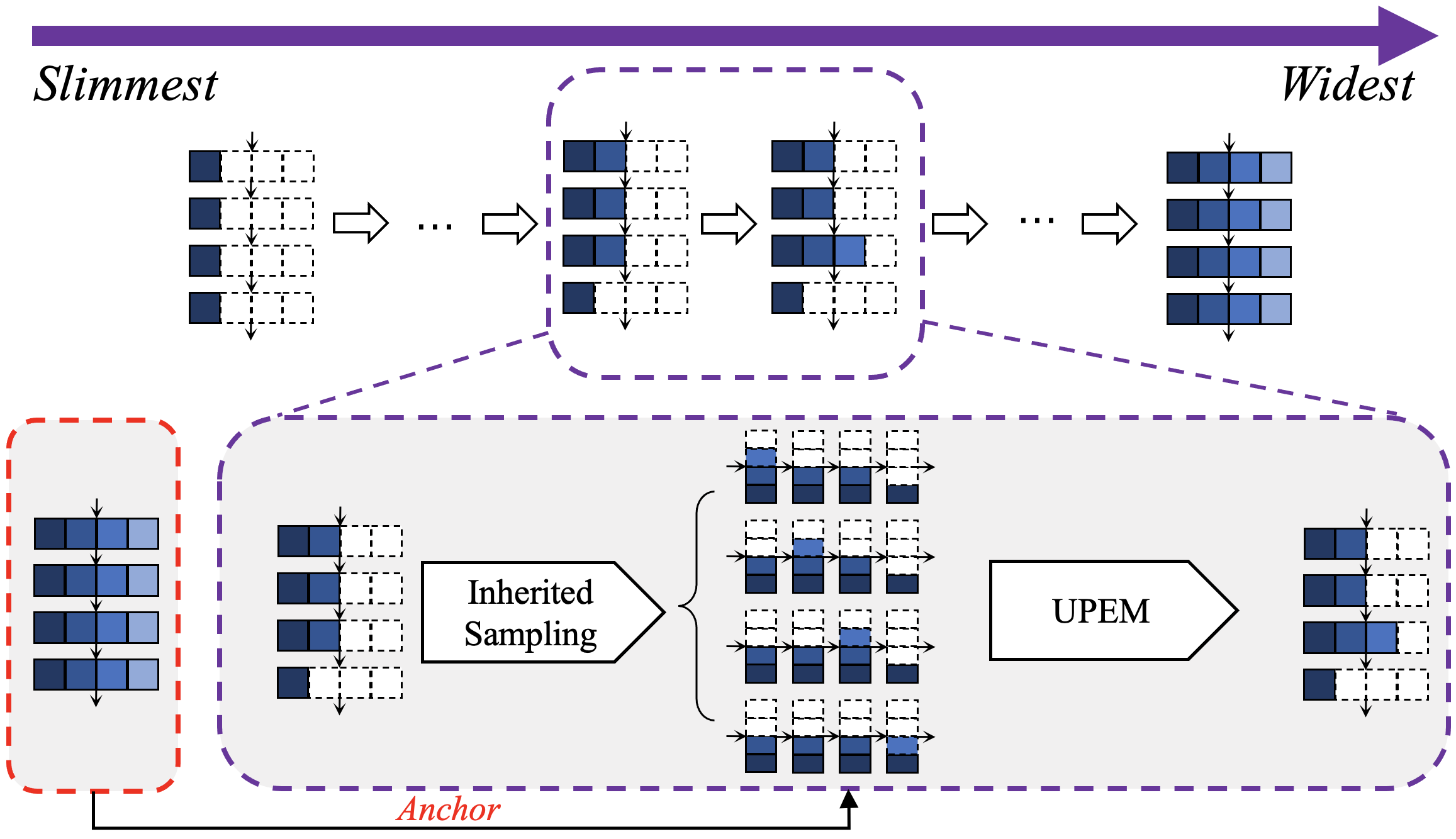}
\end{center}
   \caption{Inherited Greedy Search for channel configurations.
}
\label{searching}
\end{figure}

Although many neural architecture search methods (\emph{e.g.}, enumerate method~\cite{cai2019once} and genetic algorithms~\cite{2020Single})  can be coupled with our proposed UPEM to search optimal architectures on the unlabeled target data, we would like to provide more technique details for the efficient search method we used in the main text of the paper. As shown in Fig. \ref{searching}, we present the search method for channel configurations inspired by \cite{2020Neural}, which we dub as ``Inherited Greedy Search''. In the Inherited Greedy Search, the larger optimal models are assumed to be developed from the smaller optimal ones. Under this guidance, we obtain optimal models under different computational budgets from the slimmest to the widest models in a trip. 

\begin{algorithm}[t]
\small
\SetKwData{Left}{left}
\SetKwData{This}{this}
\SetKwData{Up}{up}
\SetKwFunction{Union}{Union}
\SetKwInOut{Input}{input}
\SetKwInOut{Output}{output}
\Input{labeled source data $\mathcal{D}_s=\{(x_i^s,y_i^s)\}_{i=1}^{n_s}$\; 
    unlabeled target data $\mathcal{D}_t=\{(x_i^t)\}_{i=1}^{n_t}$\; 
    model bank $(F,C^s,C^t,C^a)$\; learning rate $l$\; model batch size $m$\; domain confusion loss $\mathcal{L}_{dc}$\;cross-entropy loss $\mathcal{L}_{ce}$\;}
\Output{model bank $(F,C^a)$}
\BlankLine
\For{$x_i^s,x_i^t,y_i^s$ in $\mathcal{D}_s \cup \mathcal{D}_t$}
{
\emph{Stochastically Sample a Model Batch}: $\{(F_j,C^s_j,C^t_j,C^a_j)\}_{j=1}^m$\;
\emph{Calculate Model Confidence}: $\{\mathbf{g}(F_j,C^s_j,C^t_j)\}_{j=1}^m$\;
\emph{Calculate} $\{\frac{\partial \mathcal{L}_{dc}}{\partial F_j},\frac{\partial \mathcal{L}_{dc}}{\partial C^s_j},
\frac{\partial \mathcal{L}_{dc}}{\partial C^t_j}\}_{j=1}^m$\;
\emph{Generate} $\mathbf{g}_{seed}$: $\mathbf{g}_{seed}=\mathbb{E}_{\mathbf{g}(F_j,C_j^s,C_j^t)}\big[\mathbf{g}(x_i^t ; F_j,C_j^s,C_j^t)\big]$\;
\emph{Calculate} $\mathcal{L}_{seed}$:
$\mathcal{L}_{seed}= \mathcal{L}_{ce}(\mathbf{g}(x_i^t ; F_j,C_j^a), \mathbf{g}_{seed})+\mathcal{L}_{ce}(\mathbf{g}(x_i^s ; F_j,C_j^a), y_i^s)$\;
\emph{Calculate} $\{\frac{\partial \mathcal{L}_{seed}}{\partial F_j},\frac{\partial \mathcal{L}_{seed}}{\partial C_j^a}\}_{j=1}^m$\;
\emph{Update} $\{C^s, C^t, C^a\}$:
$C^s \leftarrow C^s-l\cdot\frac{1}{m}\sum_{j=1}^{m}\frac{\partial \mathcal{L}_{dc}}{\partial C^s_j}$ 
$C^t\leftarrow C^t-l\cdot\frac{1}{m}\sum_{j=1}^{m}\frac{\partial \mathcal{L}_{dc}}{\partial C^t_j}$ 
$C^a\leftarrow C^a-l\cdot\frac{1}{m}\sum_{j=1}^{m} \frac{\partial \mathcal{L}_{seed}}{\partial C_j^a}$\;
\emph{Update} $F$:
$F \leftarrow F-l \cdot\mathbb{E}_{\mathbf{g}(F_j,C_j^s,C_j^t)}\big[\frac{\partial \mathcal{L}_{dc}}{\partial F_j}\big] -l \cdot\mathbb{E}_{1-\mathbf{g}(F_j,C_j^s,C_j^t)}\big[\frac{\partial \mathcal{L}_{seed}}{\partial F_j}\big]$\;
}
\caption{Stochastic EnsEmble Distillation}
\label{algo:seed}
\end{algorithm}

\begin{figure*}[h]
\begin{center}
\includegraphics[width=.94\linewidth]{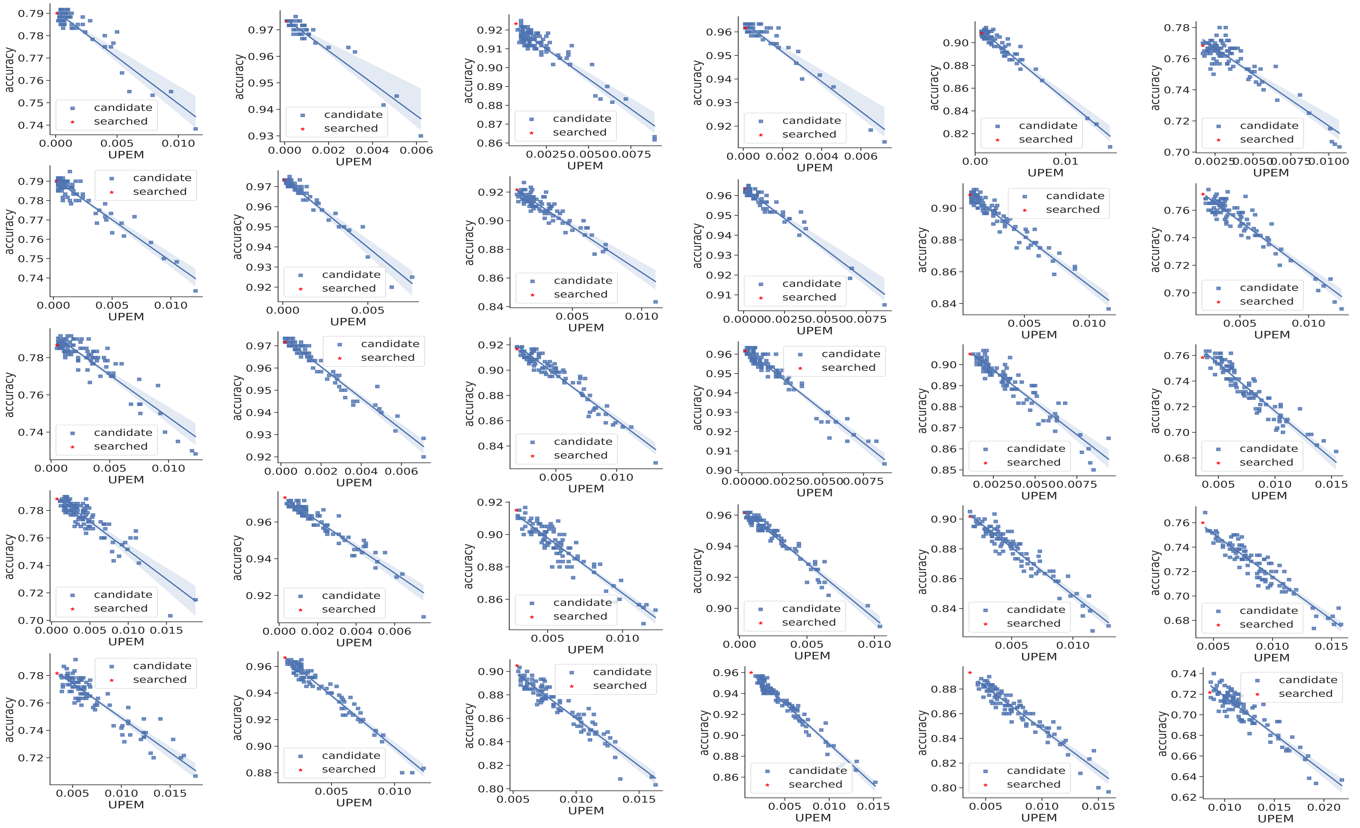}
\end{center}
\vspace{-4mm}
   \caption{The visualization of the relation of numerous models with different computational complexities on different domain adaptation tasks. Here we present the results of six adaptation tasks (six columns) on the ImageCLEF-DA dataset and five computational budgets (five rows) from $1/2\times$ to $1/32\times$. Each column from left to right presents adaptation task I$\to$P, I$\to$C, P$\to$I, P$\to$C, C$\to$I, and C$\to$P, respectively, while each row presents one computational budget on six adaptation tasks. In each sub-figure, we present the accuracy relation of 100 randomly-sampled models, and the \textcolor{red}{red} dot denotes our searched one, which is superior to most of the candidate models. 
}
\label{upem}
\end{figure*}

\begin{table*}[h]
\small
	\centering
	\resizebox{1.0\textwidth}{!}{
	\begin{tabular}{c|r|c|c|c|c|c|c|c|c|c|c|c|c|>{\columncolor{mygray}}c}
	\hline
	 methods&FLOPs &plane &bcycl & bus &car &horse & knife &mcycl&person&plant&sktbrd&train&truck&mean\\
	\hline\hline
	Source only&1$\times$&55.1 &53.3& 61.9& 59.1& 80.6& 17.9& 79.7 &31.2& 81.0 &26.5 &73.5& 8.5& 52.4
\\
	Stand-alone MCD &1$\times$&86.6 &42.9& 77.6& 76.7 &69.0& 30.6& 73.8& 70.2& 68.6& 60.5& 68.6 &9.7& 61.2
\\
	\hline\hline
	SlimDA with MCD&1$\times$ &94.3&	77.9&	84.7&	66.0&	91.9&	71.0&	88.7&	75.8&	91.2&	77.8&	86.4&	36.3&	78.5
\\
	&1/2$\times$ &94.0&	77.7&	84.5&	65.5&	91.7&	70.9&	88.6&	75.6&	90.9&	77.7&	86.4&	36.2&	78.3
\\
	&1/4$\times$ &93.7 &	77.6 &	84.3 &	65.4 &	91.5 &	70.6 &	88.4	 &75.3 &	90.8 &	77.7 &	86.4 &	36.1 &	78.1
\\
&	1/8$\times$& 93.6&	77.7&	84.2&	65.5&	91.6&	70.6&	88.5&	75.5&	90.5	&77.4&	86.4&	36.1&	78.1
\\
&	1/16$\times$& 93.5&	77.5&	84.2&	65.5&	91.7&	70.6&	88.3&	75.4&	90.6&	77.5&	86.3&	36.0&	78.1
\\
&	1/32$\times$& 93.3&	77.3&	84.0&	65.1&	91.7&	70.9&	88.1&	75.1&	90.6&	77.7&	86.3&	35.6&	78.0
\\
   & 1/64$\times$ &93.5&	77.0&	83.9&	64.9&	91.6&	70.5&	88.1&	74.9&	90.6&	77.1&	85.7&	34.9&	77.7
\\
	\hline
	\end{tabular}
	}
	\caption{Comparison among models with different FLOPs in SlimDA on VisDA datasets. We report the accuracy for each class. Moreover, ``mean'' indicates the average value among 12 per-class accuracy.}
	\label{visda-flops}
\end{table*}

\renewcommand\arraystretch{1} 
\begin{table*}[h]
    \small
	\centering
    \setlength\tabcolsep{11pt}
	\begin{tabular}{l|r|c|c|c|c|c|c |>{\columncolor{mygray}}c|>{\columncolor{mygray}}c}
	\hline
	Methods & FLOPs &I $\to$ P & P $\to$ I & I $\to$ C & C $\to$ I & C $\to$ P & P $\to$ C & Avg.&$\Delta$\\
	\hline\hline
	SlimDA&1$\times$&79.2 &92.3&97.5&91.2&76.7&96.5&88.9&--\\
&1/2$\times$  &79.0 &92.3&97.3&90.8&76.8&96.2&88.7&0.2$\downarrow$\\
&1/4$\times$  &79.0 &92.2&97.3&90.8&77.2&96.3&88.8&0.1$\downarrow$\\
&1/8$\times$  &78.7 &91.7&97.2&90.5&75.8&96.2&88.4&0.5$\downarrow$\\
&1/16$\times$  &78.8 &91.5&97.3&90.2&76.0&96.2&88.3&0.6$\downarrow$\\
&1/32$\times$ &78.2 &90.5&96.7&89.3&72.2&96.0&87.2&1.7$\downarrow$\\
&1/64$\times$ &78.3 &90.7&95.8&88.3&71.8&94.8&86.6&2.3$\downarrow$\\
    \hline\hline
	Inplaced Distillation  &1$\times$ &78.0&87.8&94.2&90.3&75.7&91.8&86.3&-\\
	&1/2$\times$ &77.1&87.0&94.0&89.0&74.0&90.9&85.3&$1.0\downarrow$\\
	&1/4$\times$ &76.3&86.3&93.5&87.7&72.6&89.7&84.3&2.0$\downarrow$\\
	&1/8$\times$ &75.5&84.8&92.9&85.5&70.9&89.3&82.7&3.6$\downarrow$\\
	&1/16$\times$ &73.3&82.6&91.5&83.9&68.0&87.4&81.2&5.1$\downarrow$\\
	&1/32$\times$ &71.1&81.0&90.5&82.2&66.5&86.5&79.6&6.7$\downarrow$\\
    &1/64$\times$ & 70.0&80.3&90.2&81.5&65.8&86.2& 79.0&7.3$\downarrow$ \\
	\hline
	\end{tabular}
	\caption{Comparison with Inplaced Distillation on ImageCLEF-DA dataset. The model batch size is set to 10 for both methods. ``$\Delta$'' means the averaged performance gap between 1$\times$ FLOPs of ResNet-50 and other models, respectively. Note that ``$\Delta$'' is calculated for SlimDA and Inplaced Distillation, respectively.}
	\label{distillation:imageclef-da}
\end{table*}

\begin{table}[h]
\small
	\centering
	\resizebox{.47\textwidth}{!}{
	\begin{tabular}{c|c||c|c}
	\hline
	 Method& mean acc.&Method &mean acc.\\
	 \hline\hline
	 ResNet-50 \cite{liang2020we}&52.1 & CDAN \cite{2017Conditional} &70.0\\
	 DANN \cite{ganin2015unsupervised} & 57.4& MDD \cite{2019Bridging}& 74.6\\
	 DAN \cite{2018Transferable} & 61.6& GVB-GD \cite{2020Gradually}&75.3\\
	 MCD \cite{2018Maximum}& 69.2& SHOT \cite{liang2020we}& 76.7\\
	 GTA \cite{2017Generate} &69.5& SHOT++ \cite{2020Source}& 77.2\\
	 1$\times$ FLOPs in SlimDA&78.5&1/64$\times$ FLOPS in SlimDA&77.7\\
	\hline
	\end{tabular}
	}
	\caption{Comparison with different UDA methods on the VisDA dataset. Note that the network backbone is set ResNet-50 in this table. ``1$\times$ FLOPs in SlimDA'' and ``1/64$\times$ FLOPs in SlimDA'' mean the models with 1$\times$ and 1/64$\times$ FLOPs of ResNet-50 in SlimDA, respectively. ``mean acc.'' means the averaged accuracy among 12 per-class accuracies.}
	\label{visda-compare}
\end{table}

Specifically, we begin with the slimmest model (the 1/64$\times$ FLOPs model) and set it as our initial model for searching. Then we divide the FLOPs gap ($F$) between the slimmest and the widest models into $k$ parts equally. 
In this way, we can search the optimal models under different computational budgets from the slimmest and the widest models in $k$-$1$ steps with steady FLOPs growing.
In the first step, we sample $q$ models by randomly adding the channels to each block of the initial model to fit the incremental FLOPs ($F/k$), and then we leverage UPEM to search the optimal model among $q$ models in this step. In the next step, we use the previous searched model as the initial model to repeat the same process in the first step, and we call this sampling method ``Inherited Sampling''. Benefited from the Inherited Greedy Search, we can significantly reduce the complexity of the search process.

\section{Additional Experiments and Analysis}
\subsection{Additional Experiments on VisDA}
We also report the results of our SlimDA on the large-scale domain adaptation benchmark, VisDA \cite{2017VisDA}, to evaluate the effectiveness of our SlimDA. VisDA is a challenging large-scale image classification benchmark for UDA. It consists of 152k synthetic images rendered by the 3D model with annotations as source data and 55k real images without annotations from MS-COCO as target data. There are 12 categories in VisDA. There is a large domain discrepancy between the synthetic style and real style in VisDA. We follow the network configuration and implementation details from the experiments on ImageCLEF-DA, Office-31, and Office-Home as described in the main text of the paper.

As shown in Table \ref{visda-flops}, our SlimDA can simultaneously boost the adaptation performance of models with different FLOPs. It can be observed that the model with 1/64$\times$ FLOPs of ResNet-50 merely has a drop of 0.8\% on the mean accuracy.
In addition, from Table \ref{visda-compare}, we can observe that our SlimDA with only 1/64$\times$ FLOPs can surpass other SOTA UDA methods by a large margin.
Overall, our SlimDA can achieve impressive performance for the large-scale dataset with significant domain shifts such as VisDA.

\subsection{Additional Analysis}

\noindent\textbf{Comparison with Inplaced Distillation.~}
Inplaced Distillation is an improving technique provided in slimmable neural network. With Inplaced Distillation, the largest model generates soft labels to guide the learning of the remaining models in the model batch. However, it cannot remedy the issue of uncertainty brought by domain shift and unlabeled data. We combine Inplaced Distillation with SymNet straightforwardly as another baseline. As shown in Table \ref{distillation:imageclef-da}, we can observe that Inplaced Distillation leads to negative transfer for the model with 1$\times$ FLOPS. Moreover, the performance of models with fewer FLOPs is limited with Inplaced Distillation. Compared with Inplaced Distillation, our SlimDA can bring performance improvements consistently under different computational budgets.
In addition, the smallest model with 1/64$\times$ FLOPs in SlimDA surpasses the corresponding counterpart in Inplace Distillation by 7.6\% on the averaged accuracy. Furthermore, even comparing the smallest model in our SlimDA to the largest one in Inplaced Distillation, the model can still achieve improvement of 0.3\% on the averaged accuracy.\\

\noindent\textbf{Additional analysis for UPEM.~}We provide additional analysis for the effectiveness of our proposed UPEM. As shown in Fig.\ref{upem}, we visualize the relation between accuracy and UPEM with different adaptation tasks and computational budgets on the ImageCLEF-DA dataset.   
The accuracy of 100 models with different network configurations under the same computational budgets vary a lot, which indicates the necessity for architecture adaptation. Moverover, the UPEM will have strict negative correlation with the accuracy for different adaptation tasks and different computational budgets (the blue lines in 30 sub-figures). Specifically, the models with the smallest UPEM (the red dots in 30 sub-figures) always have the top accuracy among models with the same computational budgets.\\

\begin{table}[t]
\large
	\centering
	\resizebox{.47\textwidth}{!}{
	\begin{tabular}{c|c|c|c|c|c|c|>{\columncolor{mygray}}c}
	\hline
	Methods &I $\to$ P & P $\to$ I & I $\to$ C & C $\to$ I & C $\to$ P & P $\to$ C& Overall\\
	\hline\hline
	stand alone 1$\times$ &0.11H&0.11H&0.11H&0.11H&0.11H&0.11H&0.66H\\
	SlimDA &0.63H&0.67H&0.60H&0.65H&0.63H&0.67H&3.85H\\
	
	\hline
	\end{tabular}
	}
	\caption{Analysis for training time on ImageCLEF-DA. We report the number of hours on one V100 GPU for training the stand-alone ResNet-50 model and our SlimDA framework with SymNet. ``Overall '' means the total training time for 6 adaptation tasks.}
	\label{train-time:imageclef-da}
\end{table}
\noindent\textbf{Analysis for Time Complexity of SlimDA.~}As shown in Table \ref{train-time:imageclef-da}, our SlimDA framework only takes about 6$\times$ the time to train a ResNet-50 model alone on imageCLEF-DA dataset. But SlimDA can improve the performance of numerous models impressively at the same time. We report the training time with one NVIDIA V100 GPU.\\

\noindent\textbf{Analysis for Architecture Configurations.~}
We provide details of channel configurations for models with different FLOPs in our SlimDA. As shown in Table \ref{arch-ip-2}-\ref{arch-ip-1/32} (in the last page), we report 30 models with the smallest UPEM for I$\to$P adaptation task under each FLOPs reduction. Note that the reported models with different architecture configurations are trained simultaneously in SlimDA with 6$\times$ the time to train a ResNet-50 model alone. Also, we sample these models from the model bank without re-training. Otherwise, it is inconceivably expensive and time-consuming if these model are trained individually. The numerous models and their impressive performances reflect that SlimDA significantly remedies the issue of real-world UDA, which is ignored by previous work.\\

\begin{table}[t]
    \large
	\centering
	\resizebox{.3\textwidth}{!}{
	\begin{tabular}{c|c|c|c|c}
	\hline
	$s$& 1$\times$ & 1/4$\times$ & 1/16$\times$ & 1/64$\times$ \\
	\hline\hline
    $1.0$ &88.3&88.2&87.5&85.7 \\
    $0.5$&88.3&88.2&87.9&86.1\\
    $0.1$&\textbf{89.0}&\textbf{88.9}&\textbf{88.4}&86.5\\
	$0.0$&88.9&88.8&88.3&\textbf{86.6} \\
	\hline
	\end{tabular}
	}
	\caption{Ablation study of $s$ in Eq.\ref{eq:general-model-confidence}.}
\end{table}
\noindent\textbf{A More General Method to Determine Model Confidence:~} A more general formula of Eq.\ref{eq:model-confidence}: 
\begin{equation}
    \begin{aligned}
    \mathbf{g}_j&=0.5{\rm sign}(a_j)|a_j|^{s}+0.5\\
    a_j&=2r_j-1
    \end{aligned}
    \label{eq:general-model-confidence}
\end{equation}
where ${\rm sign(\cdot)}$ is a sign function to produce +1 or -1, and $|\cdot|$ is to produce an absolute value. If $s\rightarrow 0$ or $s\rightarrow 1$, this formula will be specified as Eq.\ref{eq:model-confidence} in the paper or $\mathbf{g}_j=r_j/\sum_{j'}r_{j'}$, respectively. The performances tend to degrade a bit when $s\rightarrow 1$ due to the increasing weights of the small models for intra-model adaptation and knowledge ensemble. We find we can roughly set the model confidence in a hard way as defined in Eq.\ref{eq:model-confidence} in the paper to achieve considerable performance. \\

\begin{table*}[t]
\small
	\centering
	\resizebox{1.0\textwidth}{!}{
	\begin{tabular}{cccc|c||cccc|c||cccc|c}
	\hline
	block1&block2&block3&block4&acc.&block1&block2&block3&block4&acc.&block1&block2&block3&block4&acc.\\
	\hline\hline
	210&339&573&1370&79.0&100&341&559&2025&78.7&39&167&774&2048&78.8\\96&305&913&1563&78.8&168&201&976&1254&78.8&138&149&725&2044&78.8\\147&446&775&931&79.0&139&541&469&1017&78.7&95&480&538&1560&78.7\\87&236&1177&966&78.8&95&231&1193&852&78.7&78&458&588&1649&78.7\\46&380&860&1564&78.8&193&357&821&849&78.8&79&556&529&1176&79.0\\125&260&507&2048&78.8&108&233&1115&1119&79.0&125&228&671&2048&79.2\\81&544&707&830&78.8&262&319&527&846&79.2&196&392&632&1151&79.0\\125&332&819&1550&79.0&176&336&869&1011&79.0&299&214&261&1063&78.5\\82&414&685&1674&78.8&239&354&489&1111&79.2&230&228&572&1527&78.8\\118&353&641&1822&78.7&75&159&1149&1346&78.8&64&442&680&1616&79.0\\
	\hline
	\end{tabular}
	}
	\caption{Architecture configurations for the I$\to$P adaptation task on ImageCLEF-DA. We report the architecture configurations for models with 1/2$\times$ FLOPs of ResNet-50 in SlimDA. We report 30 models with the smallest UPEM. [``block1''$\to$ ``block4''] represents the channel number for each block consisting of layers with the same spatial resolution of feature maps.}
	\vspace{-3mm}
	\label{arch-ip-2}
\end{table*}
\begin{table*}[t]
\small
	\centering
	\resizebox{1.0\textwidth}{!}{
	\begin{tabular}{cccc|c||cccc|c||cccc|c}
	\hline
	block1&block2&block3&block4&acc.&block1&block2&block3&block4&acc.&block1&block2&block3&block4&acc.\\
	\hline\hline
	74&241&598&970&79.0&38&242&489&1293&79.0&101&98&779&654&78.5\\92&113&415&1492&78.7&134&222&353&1099&78.8&113&250&252&1260&79.2\\95&137&472&1373&79.0&114&129&307&1479&78.8&90&249&577&902&78.8\\63&130&566&1361&79.2&92&152&254&1578&78.5&134&149&557&951&79.0\\182&149&308&898&79.0&49&261&643&873&79.0&48&319&404&1112&79.0\\142&207&514&790&78.8&104&354&297&775&78.8&154&192&208&1174&79.0\\63&130&555&1380&79.2&86&200&666&876&79.0&67&334&440&899&78.8\\180&141&359&869&78.8&95&195&705&692&79.0&63&136&777&834&78.8\\87&195&736&628&79.0&139&272&459&633&78.8&115&110&662&926&79.0\\45&367&450&754&78.8&87&323&427&878&78.8&149&109&660&547&79.0\\

	\hline
	\end{tabular}
	}
	\caption{Architecture configurations for the I$\to$P adaptation task on ImageCLEF-DA. We report the architecture configurations for models with 1/4$\times$ FLOPs of ResNet-50 in SlimDA. We report 30 models with the smallest UPEM.}
	\vspace{-3mm}
	\label{arch-ip-1/4}
\end{table*}
\begin{table*}[t]
\small
	\centering
	\resizebox{.96\textwidth}{!}{
	\begin{tabular}{cccc|c||cccc|c||cccc|c}
	\hline
	block1&block2&block3&block4&acc.&block1&block2&block3&block4&acc.&block1&block2&block3&block4&acc.\\
	\hline\hline
	52&142&387&685&78.7&42&189&325&673&78.5&66&86&286&915&78.7\\67&140&385&605&79.0&93&122&316&615&78.8&90&126&350&549&78.5\\52&160&367&667&78.8&73&171&302&628&78.8&112&100&244&602&78.5\\48&173&399&554&78.7&81&154&381&425&79.2&45&90&497&564&78.8\\41&150&418&631&78.7&63&124&412&624&78.7&48&85&486&600&78.7\\52&117&504&401&78.8&108&79&286&624&78.5&92&115&280&703&78.5\\58&161&177&899&79.0&39&81&485&650&78.8&68&102&179&987&78.3\\55&83&492&541&78.7&57&198&333&533&79.0&69&202&341&375&78.7\\41&70&413&835&78.3&69&169&410&367&78.0&39&228&199&690&78.8\\43&86&225&1063&78.2&101&83&330&599&78.3&71&223&198&544&78.7\\
	\hline
	\end{tabular}
	}
	\caption{Architecture configurations for the I$\to$P adaptation task on ImageCLEF-DA.  We report the architecture configurations for models with 1/8$\times$ FLOPs of ResNet-50 in SlimDA. We report 30 models with the smallest UPEM.}
	\vspace{-3mm}
	\label{arch-ip-1/8}
\end{table*}
\begin{table*}[t]
\small
	\centering
	\resizebox{.96\textwidth}{!}{
	\begin{tabular}{cccc|c||cccc|c||cccc|c}
	\hline
	block1&block2&block3&block4&acc.&block1&block2&block3&block4&acc.&block1&block2&block3&block4&acc.\\
	\hline\hline
	41&83&136&477&78.2&45&88&138&435&78.0&37&82&241&298&77.8\\49&85&143&407&76.8&32&77&163&505&77.3&35&73&157&507&77.2\\36&132&139&284&78.7&32&64&284&257&78.5&53&77&140&406&77.8\\53&99&151&297&78.0&36&136&134&268&77.8&33&113&192&311&78.0\\53&103&150&279&78.2&32&64&132&564&77.3&39&85&198&391&77.3\\48&95&194&273&77.8&51&74&183&360&77.7&53&81&189&297&76.8\\46&89&140&423&77.8&46&97&173&338&76.8&40&132&130&267&78.2\\41&64&128&529&77.2&59&72&142&366&77.5&49&103&135&350&77.7\\42&82&134&480&77.5&40&92&213&323&77.8&37&68&152&517&77.2\\54&65&178&374&76.0&34&128&137&328&78.3&65&81&128&284&79.2\\
	\hline
	\end{tabular}
	}
	\caption{Architecture configurations for the I$\to$P adaptation task on ImageCLEF-DA. We report the architecture configurations for models with 1/32$\times$ FLOPs of ResNet-50 in SlimDA. We report 30 models with the smallest UPEM.}
	\vspace{-3mm}
	\label{arch-ip-1/32}
\end{table*}
\end{document}